\documentclass{article}

\usepackage{arxiv}
\usepackage[utf8]{inputenc}
\usepackage[T1]{fontenc}
\usepackage{url}
\usepackage{float}
\usepackage{booktabs}
\usepackage{enumitem}
\usepackage{amssymb}
\usepackage{amsthm}
\usepackage{graphicx}
\usepackage{placeins}
\usepackage{wrapfig}
\usepackage{microtype}
\usepackage{natbib}
\usepackage{hyperref}

\usepackage{amsmath,amsfonts,bm}

\def\eqref#1{equation~\ref{#1}}
\def\1{\bm{1}}

\DeclareMathAlphabet{\mathsfit}{\encodingdefault}{\sfdefault}{m}{sl}
\SetMathAlphabet{\mathsfit}{bold}{\encodingdefault}{\sfdefault}{bx}{n}

\newtheorem{proposition}{Proposition}

\title{When the Wrong Key Wins: Understanding and Detecting Hallucinations in LLMs}

\author{
\normalfont
Xuhan Tong$^{1}$\quad
Haoyue Bai$^{1}$\quad
Dawei Zhou$^{2}$\\
Naichen Shi$^{3}$\quad
Jiawei Zhang$^{1}$\\[3pt]
$^{1}$University of Wisconsin--Madison\\
$^{2}$Virginia Polytechnic Institute and State University\\
$^{3}$Northwestern University
}

\date{}

\begin{document}

\maketitle

\begin{abstract}
We introduce a latent-key account of why large language models can hallucinate even when the knowledge required for a correct 
answer is already available. In this account, answer selection depends on competition among associations acquired during pretraining. 
Our results provide evidence consistent with this account: model predictions can be highly sensitive to individual query keywords, 
these influential keywords exhibit entity-specific binding, and their effects vary 
systematically with pretraining frequency. Multiple bindings can also compete and
exhibit higher-order interactions within the same query.
Motivated by this account, we introduce a two-stage keyword-perturbation method for 
hallucination detection. By removing influential keywords and measuring how the model 
reorganizes its prediction, the method distinguishes errors associated with misleading key 
associations from correct decisions supported by diagnostic evidence. Across multiple 
models and benchmarks, perturbation provides a strong and transferable detection signal,
reaching $0.910$ AUROC on probe-known ScientistQA.
Finally, we extend the probabilistic framework to four hallucination regimes:
knowledge deficit, wrong knowledge, context distraction, and unstable inference. Their
distributions across benchmarks help explain why different detector families succeed in
different settings.
\end{abstract}

% !TeX root = arxiv.tex

\section{Introduction}
\label{sec:introduction}
Large language models (LLMs) have achieved strong performance across a broad range
of tasks \citep{openai2024gpt4,geminiteam2025gemini}, but capability does not guarantee 
factual reliability.  TruthfulQA shows that models can reproduce common human falsehoods 
\citep{lin2022truthfulqa}, while HaluEval documents hallucinated content across 
knowledge-intensive question answering, dialogue, and summarization \citep{li2023halueval}.  
This problem becomes more consequential as LLMs are embedded in tool-augmented and agentic 
workflows, including web search and external-service interaction \citep{nakano2022webgpt,liu2023webglm}.  
Reliably detecting hallucinations is therefore essential before model outputs trigger real-world actions.

One particularly difficult case is a factually incorrect answer produced with high confidence 
even though the required knowledge is behaviorally accessible.  Existing detectors provide 
limited insight into such errors.  Uncertainty-based methods use likelihood or self-evaluation across samples \citep{kadavath2022mostlyknow,manakul2023selfcheckgpt,farquhar2024semanticentropy}, 
but systematic errors can be confident and repeatable.  Hidden-state and attention-based 
detectors identify internal trajectories associated with incorrectness \citep{chen2024inside,du2024haloscope,chuang2024lookback,binkowski2025spectral}, 
but a confidently reasoned wrong answer may follow a trajectory that is difficult to distinguish
from that of a correct answer \citep{janiak2025illusion}.  Evidence-grounded 
methods verify claims against retrieved sources \citep{min2023factscore}, but depend 
on external evidence rather than the cause of the model's decision.  None 
of these approaches reveals which input evidence drives the mistake.

Prior work relates these errors to conflicts between statistical associations and
prompt-supported reasoning \citep{mckenna2023sources,sun2025subsequence}. However,
these studies do not identify which keyword activates the competing association in
an individual query or test how intervening on that keyword directionally changes
answer selection. Most directly, \citet{hu2026reasoningpriors} formalize the
phenomenon as inference misalignment and introduce
ScientistQA to distinguish missing knowledge from failures to deploy accessible
facts. Their analysis establishes the theoretical setting and demonstrates the
phenomenon behaviorally, but leaves unresolved how competing associations are
expressed and selected within a particular prompt.

We introduce a latent-key view of inference and targeted keyword perturbations
to study these questions. Our results provide evidence consistent with an
account in which model predictions depend on entity--keyword bindings shaped by
pretraining exposure and multiple bindings can compete within the same query.
This motivates a
two-stage perturbation detector (Figure \ref{fig:2stage-perturb}): we first identify an influential association,
then remove it and measure how the remaining evidence reorganizes the model's
prediction. The resulting perturbation responses capture how suppressing an
influential key changes the evidence supporting the model's answer. Across
multiple models and benchmarks, these responses provide a strong and
transferable signal for both hallucination detection and attribution.

Finally, we extend the same probabilistic framework beyond context-induced
misrouting. We distinguish four broad hallucination regimes: knowledge-deficit
errors, wrong-knowledge errors, context-distraction errors, and
unstable-inference errors. These regimes correspond to different points of
failure in the inference process and favor different diagnostic signals.
Perturbation is particularly informative for context distraction, while
uncertainty-based methods are better suited to unstable inference and
evidence-based methods are most useful when the model's underlying knowledge
itself is unreliable. Their complementary behavior helps explain why no single
hallucination detector performs uniformly well across benchmarks. Our main contributions are summarized as follows:
\begin{figure}[t]
    \centering
    \includegraphics[width=0.97\textwidth]{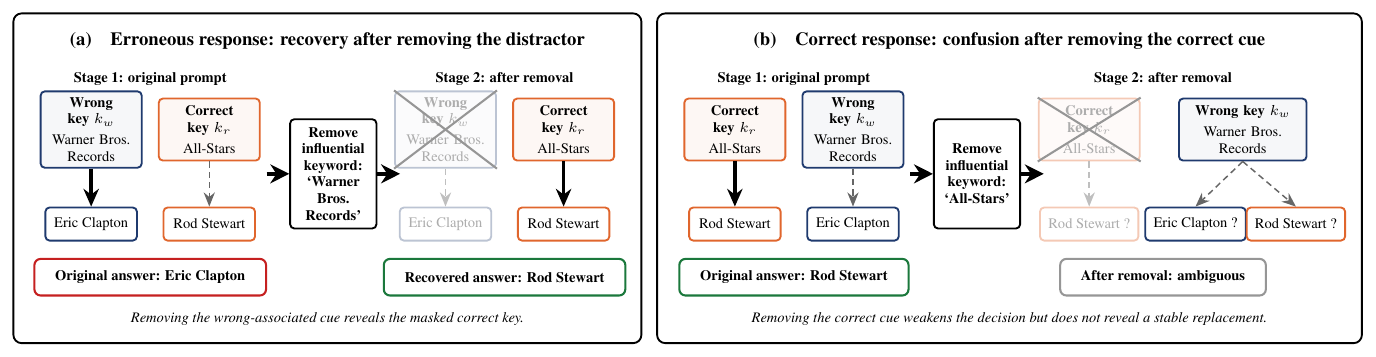}
    \caption{\textbf{Overview of two-stage keyword-perturbation detection}. Starting from
     the model's original response, we identify and remove the keyword that most 
     strongly affects its relative preference between the two candidate answers, 
     and then regenerate the answer from the perturbed prompt. \textbf{(a)}: an erroneous 
     response exhibits strong dependence on a misleading keyword association; removing that keyword 
     can suppress the wrong key and increase support for the correct answer. \textbf{(b)}: a correct 
     response relies on a diagnostic keyword; removing it weakens the correct key 
     and leaves the model ambiguous rather than systematically favoring the alternative.}
    \label{fig:2stage-perturb}
\end{figure}

\begin{itemize}[leftmargin=*,labelsep=2pt]
\item \textbf{We provide evidence consistent with a latent-key/entity-binding account of hallucination even when the relevant knowledge has
been acquired (Section~\ref{sec:binding}).} Model predictions can depend strongly on particular
keywords that are bound to question entities, and that these entity--keyword
bindings systematically influence answer selection. Controlled experiments further
provide evidence that binding strength varies with frequency imbalance during
pretraining. We also characterize interactions among multiple bindings, including
competitive and higher-order effects.

\item \textbf{We introduce a two-stage keyword-perturbation method for hallucination
detection (Section~\ref{sec:detection}).} The method first identifies influential associations and then removes them to measure how the
model reorganizes its prediction. Erroneous responses often exhibit a characteristic recovery trajectory in which suppressing a misleading key reveals a previously masked correct key, whereas
perturbing an originally correct response more often leads to ambiguity rather than systematic recovery. This counterfactual trajectory provides both a detection signal and an interpretable view of
the evidence driving the model’s decision.

\item \textbf{We develop a taxonomy linking hallucination categories to suitable 
detectors (Section~\ref{sec:category}).} We distinguish knowledge-deficit errors, wrong-knowledge
errors, context-distraction errors, and unstable-inference errors, and characterize
their empirical signatures across benchmarks. This framework explains why no
single detector family performs best universally and why complementary signals can
be combined.
\end{itemize}

\section{Entity--Keyword Binding Effects}
\label{sec:binding}

As a motivating example, consider a prompt asking the model to choose between
Eric Clapton and Rod Stewart:

\begin{quote}
``This musician is associated with the record label Warner Bros. Records.
However, this musician was not associated with the act All-Stars. Who is this
person?''
\end{quote}

Eric Clapton was associated with the All-Stars, whereas Rod Stewart was not;
the explicit negative constraint therefore rules out Clapton. Nevertheless,
the model selects Eric Clapton. One possible explanation is that the query
activates several associations simultaneously. Although ``Warner Bros.
Records'' is compatible with both musicians, its association with Clapton may
be substantially stronger in pretraining, causing the model to retrieve the
Clapton-related association even when another part of the prompt supports
Stewart.

We evaluate this account progressively. We first formulate answer selection as
competition between latent keys and show that individual query phrases can
strongly alter candidate preference. We then test whether this sensitivity
reflects entity-specific keyword bindings. Next, we manipulate pretraining
frequency to test whether the prior strength of competing keys causally affects
answer selection. Finally, we examine how several bindings interact when they
are activated simultaneously.

\subsection{Latent Key Model}
\label{sec:latent-key}

We adopt the latent-key view of \citet{hu2026reasoningpriors}. Let $x_q$ denote
the query and $z$ the remaining prompt context. We consider two competing
latent keys: $k_r$, which retrieves the relation supporting the correct answer
$y_r$, and $k_w$, which retrieves a competing relation supporting the wrong
answer $y_w$. The model prediction marginalizes over the latent key,
\begin{equation}
P(y\mid z,x_q)
=
\sum_k P(k\mid z,x_q)P(y\mid z,x_q,k).
\label{eq:latent-key-mixture}
\end{equation}
Our setting focuses on cases in which the relevant knowledge is available once
the appropriate key has been selected, $P(y_r\mid z,x_q,k_r)\approx 1$ and $P(y_w\mid z,x_q,k_w)\approx 1$,
with little probability assigned to the opposite candidate under each key.
Under this separation, the competition between the two answers is governed
primarily by the competition between their corresponding keys.

The remaining question is therefore why a query activates one key more strongly
than another. Let
$P_{\mathrm{pre}}$ denote the pretraining distribution and define the prior
probability of a key as $\pi(k)=P_{\mathrm{pre}}(k)$.
\begin{proposition}[Key-odds decomposition]
\label{prop:key-odds-decomposition}
Assume that the fixed surrounding context $z$ does not differentially reweight
the competing keys and that query--key likelihood ratios inherit the
corresponding pretraining co-occurrence statistics. Under these conditions,
the query-dependent contribution to key selection is proportional to the
relative pretraining association between the query and the competing keys:
\begin{equation}
\frac{P(y_r\mid z,x_q)}
     {P(y_w\mid z,x_q)}
\propto
\frac{P_{\mathrm{pre}}(k_r,x_q)}
     {P_{\mathrm{pre}}(k_w,x_q)}
=
\frac{\pi(k_r)}
     {\pi(k_w)}
\frac{P_{\mathrm{pre}}(x_q\mid k_r)}
     {P_{\mathrm{pre}}(x_q\mid k_w)}.
\label{eq:pretraining-key-odds}
\end{equation}
\end{proposition}
This proposition is a conditional model: Bayes' rule yields the posterior-odds
decomposition, while the connection to pretraining statistics follows from the
two assumptions stated above rather than from Bayes' rule alone.
Proposition~\ref{prop:key-odds-decomposition} gives a simple interpretation of
answer selection. Once each key reliably retrieves its associated answer, the
competition between $y_r$ and $y_w$ is governed primarily by which key receives
greater weight. This preference has two components: the prior
$\pi(k)$ captures the overall strength of a key acquired during pretraining,
while $P_{\mathrm{pre}}(x_q\mid k)$ captures how strongly the current query is
associated with that key. A wrong answer can therefore arise when the wrong key
is favored by its prior, by its binding to the current query, or by both.

Although these quantities are latent, the decomposition suggests a sequence of
testable consequences. First, if the query--key association is concentrated in
a small number of informative phrases, neutralizing such a phrase should
produce a large change in answer preference, whereas perturbing a matched
weakly associated span should have little effect. This establishes that answer
selection is locally sensitive to particular query features. Second, local
sensitivity alone does not establish binding: an influential phrase could
simply be generically informative. If its effect is induced by an
entity--keyword association, the same keyword should also be preferentially
retrieved when conditioning on its associated entity rather than on the
competing entity. Finally, the prior term predicts where these
preferences can originate: increasing the pretraining frequency of one
entity--keyword association, while keeping the evaluation query structure
fixed, should shift answer preference toward the corresponding
key. Section~\ref{sec:empirical} tests these predictions in this order. A formal derivation of
Proposition~\ref{prop:key-odds-decomposition} from the latent key--task
framework is provided in Appendix~\ref{app:key-odds-proof}.

\subsection{Empirical Verification}
\label{sec:empirical}
Section~\ref{sec:latent-key} shows that, once competing keys reliably retrieve
their corresponding answers, answer selection depends jointly on the keys'
pretraining priors and their associations with the current query.
In this subsection we test the three predictions of Equation~\ref{eq:pretraining-key-odds} in turn.
\paragraph{I. Local Sensitivity}
The first prediction concerns the query-dependent association term in
Equation~\ref{eq:pretraining-key-odds}. If this association is concentrated in
a small number of attribute keywords, neutralizing one of them should produce
a large change in the relative preference between $y_r$ and $y_w$. To measure this dependence,
we score each candidate by its length-normalized conditional log-likelihood
$\ell(y\mid x)$ for the full prompt $x=(z,x_q)$ and define the wrong--right
margin $m(x)
=
\ell(y_w\mid x)-\ell(y_r\mid x)$.
For a query keyword $k$, let $x\setminus k$ denote the input obtained by
replacing the embeddings of $k$ with the model-wide mean embedding. We measure
its perturbation effect by $u_k=m(x)-m(x\setminus k)$.

\begin{table}[t]
\centering
\caption{Keyword perturbation effects and owner-aligned frequency differences for erroneous and correct responses. Perturbation denotes the change in the wrong-versus-right answer margin after neutralizing the attribute keyword. Complete results and corresponding control values are reported in Appendix~\ref{app:binding-empirical}.}
\label{tab:keyword_perturbation_binding}
\small
\setlength{\tabcolsep}{4pt}
\begin{tabular}{lccc|ccc}
\toprule
& \multicolumn{3}{c}{\textbf{Erroneous responses}}
& \multicolumn{3}{c}{\textbf{Correct responses}} \\
\cmidrule(lr){2-4}\cmidrule(lr){5-7}
Attribute
& $n$ & $\lvert u_k\rvert$ & $\lvert\Delta f\rvert$
& $n$ & $\lvert u_k\rvert$ & $\lvert\Delta f\rvert$ \\
\midrule
\textbf{Overall}
& \textbf{454}
& $\mathbf{0.453}$
& $\mathbf{0.337}$
& \textbf{612}
& $\mathbf{0.500}$
& $\mathbf{0.319}$ \\
Award received
& 324
& $0.427$
& $0.488$
& 374
& $0.438$
& $0.447$ \\
Education
& 144
& $0.648$
& $0.287$
& 221
& $0.723$
& $0.337$ \\
Field
& 192
& $0.139$
& $0.191$
& 263
& $0.225$
& $0.212$ \\
Occupation
& 79
& $0.104$
& $0.081$
& 148
& $0.215$
& $0.146$ \\
Position held
& 63
& $0.556$
& $0.212$
& 96
& $0.636$
& $0.213$ \\
\bottomrule
\end{tabular}
\end{table}

We test local sensitivity on ScientistQA by matching query keywords to five
profile attributes: award received, education, field, occupation, and position
held. For each matched keyword, we neutralize its embeddings while leaving all
other tokens and positions unchanged. A nearby same-width non-attribute span
serves as a control for intervention length and position. The analysis contains
454 erroneous and 612 correct responses. Because a question may contain
eligible keywords from multiple attribute families, the category-specific
$n$ values overlap and therefore sum to more than the Overall count. As shown in Table~\ref{tab:keyword_perturbation_binding}, individual attribute
keywords exert substantial control over answer preference. Thus, model decisions can be sharply altered by neutralizing a
small part of the query. This result establishes local sensitivity but does not 
show whether the keyword is preferentially bound to one entity rather than being 
globally informative. The next experiment tests this entity-specific association.

\paragraph{II. Entity--Keyword Binding}

Equation~\ref{eq:pretraining-key-odds} predicts that a query can favor one
key over another when it has been more strongly associated with that key during
pretraining. If a keyword is bound to one entity, this association
should also be observable in the reverse direction: conditioning on that
entity should make the keyword more likely to be retrieved than the competing entity.

We test this prediction with a separate free-generation assay over the same
attributes used in the perturbation experiment. We prompt the model to describe
each candidate with respect to an attribute compatible with both and measure
the difference in exact keyword-mention frequency between them:
$\Delta f
=
P(k\text{ mentioned}\mid e_{\mathrm{owner}})
-
P(k\text{ mentioned}\mid e_{\mathrm{other}})$,
where $e_{\mathrm{owner}}$ is the candidate whose profile supplies $k$ and
$e_{\mathrm{other}}$ is the competing candidate.
This behavioral quantity serves as a proxy for the relative
query--key binding term in Equation~\ref{eq:pretraining-key-odds}.

Table~\ref{tab:keyword_perturbation_binding} shows positive owner-aligned
frequency differences across all five attribute families. Averaged over
attributes, Exact $\Delta f$ is $0.337$ for erroneous responses and $0.319$
for correct responses. Thus, an attribute shared by both candidates need not
be neutral: conditioning on one candidate can retrieve its keyword more
strongly than conditioning on the other. This asymmetry gives a concrete
interpretation of the wrong keyword $k_w$ introduced in
Section~\ref{sec:latent-key}: $k_w$ is selected when a shared attribute is more
strongly bound to the incorrect candidate and this association gives it greater
weight than $k_r$ in the current query. Once $k_w$ dominates, the model can
select the wrong answer even though the attribute itself is compatible with
both candidates.

Importantly, both correct and erroneous responses exhibit strong bindings.
Hallucination is therefore not caused by the mere existence of an
entity--keyword association. In correct responses, the candidate-aligned
binding does not outweigh the query's logical evidence for $k_r$, so the model
still selects the correct answer. In erroneous responses, the binding favoring
$k_w$ instead dominates this evidence, preventing the logical constraint from
correcting the prediction. The distinction lies in the balance
between the competing signals, rather than in the absolute
magnitude of a binding alone.

\paragraph{III. Controlled Frequency Exposure Shifts Key Priors}

We next test the first factor in Equation~\ref{eq:pretraining-key-odds}: the
prior ratio $\pi(k_r)/\pi(k_w)$, which captures the relative prior strength of
the competing keys. Since this prior is acquired from the pretraining
distribution, the model predicts that increasing the frequency of one
association during pretraining should systematically increase its influence at
inference time.

\begin{table}[t]
\centering
\caption{Frequency--dose response across models. Values are candidate-preference differences $u_\Delta$; larger values indicate stronger transfer of asymmetric exposure frequency into candidate preference. Values are means across random seeds 42, 43, and 44; complete per-seed results are reported in Appendix~\ref{app:binding-empirical}.}
\label{tab:binding-dose-three-models}
\small
\setlength{\tabcolsep}{3pt}
\begin{tabular}{@{}lcccccc@{}}
\toprule
Model & $d=0$ & $d=.25$ & $d=.5$ & $d=.75$ & $d=1$ & Slope \\
\midrule
Qwen2.5
& .028 & .055 & .131 & .200 & .232 & .221 \\
\midrule
Llama-3.2
& .002 & .013 & .014 & .024 & .064 & .054 \\
\midrule
Ministral-3
& .013 & .059 & .095 & .285 & .564 & .531 \\
\bottomrule
\end{tabular}
\end{table}

We test this prediction with a controlled frequency--dose continued-training
experiment applied to the last four transformer blocks of an existing
checkpoint. We
construct 50 mirrored pairs of fictional people and facts, with each pair
containing 4 shared facts and 1 constraint fact. We use a dose parameter
$d$ to control the relative continued-training exposure of one of the shared
facts, and see how it affects the model's preference; the 
assignment is reversed for the other person. 
For each pair, we compare the candidate-preference margin under the
two competing facts. Let $x_w$ and $x_r$ denote the corresponding evaluation
prompts. We report $u_\Delta
=
m(x_w)-m(x_r)$, and fit a linear slope across the five dose levels.

Table~\ref{tab:binding-dose-three-models} shows this predicted dose-response
pattern.
As the dose increases, the mean preference difference rises
across all three models, and the fitted slopes are also positive, meaning that
increasing the controlled training exposure of an association systematically shifts
later answer preference in its direction.

This experiment provides evidence consistent with the prior term in
Equation~\ref{eq:pretraining-key-odds}. Together with the binding experiment above, the results
suggest a two-factor account in which key selection depends jointly on
\emph{how frequent a key is} and \emph{how strongly the current query is bound
to that key}.

\subsection{Interactions among Multiple Keyword Bindings}
\label{sec:multi}

Section~\ref{sec:latent-key} isolates a pair of competing keys to formalize
the central account. Real prompts, however, can activate several
entity--keyword associations simultaneously. In the motivating example,
different phrases may support different latent keys, and the effect of one
keyword may depend on whether another remains present. We therefore 
enumerate the complete perturbation factorial over the candidate
keywords in ScientistQA. This allows us to distinguish four broad pairwise
patterns. Two keywords \emph{compete} when their signed perturbation effects
push answer preference in opposite directions; \emph{synergy} and
\emph{redundancy} capture consistent non-additive interactions. We additionally report higher-order mass: the
share of non-additive interaction strength attributable to combinations of at
least three keywords, which describes where the interaction strength comes from.

Table~\ref{tab:keyword-interactions} shows that keyword effects are
not independent. Competition occurs in $19.3\%$ of keyword pairs for erroneous
responses, compared with $13.7\%$ for correct responses. Redundancy
($9.2\%$ versus $9.5\%$) and synergy ($0.6\%$ versus $0.7\%$) are nearly
unchanged, while Other accounts for $70.9\%$ and $76.1\%$ of pairs,
respectively. Higher-order interaction mass is substantial in both groups.

These results complete the empirical predictions developed in this section.
Within the conditional model, keys have different prior strengths, queries
bind asymmetrically to these keys, and several bindings can be activated
together at inference time.
First, the higher competition rate among erroneous responses supports the
central account of error: influential keywords more often favor opposing
answers, and an error occurs when their combined influence makes $k_w$
dominate $k_r$ despite the correct answer remaining available. Second, the
large Other category suggests that strong effects are concentrated
in a relatively small subset of keyword pairs, consistent with the framework's
assumption that answer selection is dominated by a few salient associations. 
Third, the substantial higher-order mass
shows that these associations can also combine in ways that cannot be reduced
to isolated or pairwise effects. Consequently, the absolute influence of one
keyword cannot by itself identify hallucination: strong bindings occur in both
correct and erroneous responses, whereas errors depend on which answers the
keywords support and how their effects combine. Section~\ref{sec:detection}
therefore uses the joint pattern of perturbation effects rather than
perturbation magnitude alone.

\begin{table}[t]
  \centering
  \caption{
  Keyword interaction structure in correct and erroneous responses.
  Pairwise percentages are computed within each question and then averaged,
  giving every question equal weight. Higher-order mass is the share of
  absolute interaction mass at order two or above attributable to combinations
  of at least three keywords. Complete results are reported in
  Appendix~\ref{app:multi}.
  }
  \label{tab:keyword-interactions}
  \small
  \setlength{\tabcolsep}{5pt}
  \begin{tabular}{lrrcccc|c}
  \toprule
  Subset
  & $n$
  & Pairs
  & Competition
  & Redundancy
  & Synergy
  & Other
  & Higher-order mass \\
  \midrule
  Correct responses
  & 612
  & 2,347
  & 13.7\%
  & 9.5\%
  & 0.7\%
  & 76.1\%
  & 28.5\% \\

  Erroneous responses
  & 454
  & 1,781
  & 19.3\%
  & 9.2\%
  & 0.6\%
  & 70.9\%
  & 29.9\% \\
  \bottomrule
  \end{tabular}

  \end{table}

\section{Hallucination Detection via Keyword Perturbation}
\label{sec:detection}
\subsection{Perturbation Rationale}
\label{sec:mechanism}
Section~\ref{sec:binding} provided evidence that both correct and erroneous responses can
depend strongly on entity--keyword bindings. A large perturbation
effect alone is therefore insufficient to identify hallucination. We therefore
examine what happens \emph{after the currently dominant association is removed}.

Let $y_c$ denote the
candidate originally selected by the model and $y_a$ the competing candidate.
For the original prompt $x$, we define the chosen-versus-alternative margin $m(x)
=
\ell(y_c\mid x)-\ell(y_a\mid x)$.
For a candidate keyword $k_i$, we first neutralize its embeddings and
measure how the current prediction changes. This first-stage intervention
identifies keywords on which the selected answer strongly depends. Importantly,
such a keyword may correspond either to $k_w$ in an erroneous response or to
$k_r$ in a correct response; first-stage sensitivity alone therefore
does not distinguish the two cases.

An informative distinction may emerge after the influential keyword is removed. Consider first
an erroneous response. Under the latent-key model of
Section~\ref{sec:latent-key}, the model answers incorrectly because a wrong
keyword $k_w$ dominates the correct keyword $k_r$, even though the knowledge associated
with $k_r$ remains available. In the motivating example,
``Warner Bros. Records'' may activate the Clapton-associated keyword strongly enough
to dominate the more diagnostic ``All-Stars'' relation. Perturbing or removing
the former suppresses $k_w$, allowing the previously masked correct keyword $k_r$
to become dominant. The model can therefore recover from $y_w$ to $y_r$.
Once the shortcut keyword has been removed, perturbing the remaining correct
keyword should again produce a strong effect, because that keyword now governs
the recovered prediction.

The behavior is different when the original response is correct. In this case,
the model already selects the appropriate keyword $k_r$. Removing the keyword that
supports $k_r$ destroys the evidence responsible for the correct prediction,
but need not reveal a comparably strong competing keyword. In the motivating
example, after removing ``All-Stars'', ``Warner Bros. Records'' may be
compatible with both candidates without functioning as a decisive keyword. The
model therefore becomes less certain about which candidate to select rather
than undergoing the systematic keyword replacement observed in the erroneous
case. Subsequent perturbations of the remaining keywords consequently produce a
weaker or more diffuse response.

This motivates a two-stage perturbation procedure. In the first stage, we
perturb each candidate keyword and measure its effect on the original
chosen-versus-alternative margin. For an influential keyword $k_i$, we then
remove $k_i$ from the prompt and repeat the perturbation analysis over the remaining
keywords. The second stage therefore measures how the model reorganizes its
evidence after one association has been removed.

This distinction is expected to be strongest when the relevant knowledge is
already available to the model. In the knowledge-known regime,
$P(y_r\mid x,k_r)$ is concentrated on the correct answer. Thus, when removing
$k_w$ causes $k_r$ to become dominant, the change in keyword selection produces an
observable shift toward $y_r$. When the knowledge is unavailable, increasing
the preference for $k_r$ need not reliably favor the correct answer, and the
resulting perturbation response can resemble the confusion induced by removing
$k_r$ from an originally correct response. We therefore expect two-stage keyword
perturbation to be particularly distinguishable for knowledge-known
hallucinations, while its signal becomes weaker as the mapping from the correct
keyword to the correct answer becomes less reliable.

\subsection{Detection Performance}
\label{sec:detection-performance}

We next test whether the two-stage perturbation signatures described above can
reliably distinguish erroneous from correct responses. For each question, we
extract features from both the original perturbations and the second-stage
perturbations after removing an influential keyword, and train a
classifier to predict whether the model's original answer is correct.
We evaluate the detector on ScientistQA, TriviaQA, GSM8K, and DROP using three
model families. ScientistQA additionally provides a probe-known subset, in
which the relevant knowledge is independently verified to be available to the
model. This subset evaluates the regime predicted by the account above:
when the correct key reliably retrieves the correct answer, removing a
misleading key should produce a particularly informative perturbation response. For the
open-ended GSM8K and DROP benchmarks, we use an LLM to generate an alternative
candidate answer so that the same paired-candidate perturbation procedure can
be applied. Appendix~\ref{app:dataset-construction} details this
candidate-construction process and evaluates a modified perturbation method
that does not require a paired alternative.

To further evaluate generalization beyond the entities and relations seen in
ScientistQA, we construct a new multi-domain extension covering
\emph{athletes}, \emph{musicians}, and \emph{buildings}. The extension follows
the same paired-candidate construction as ScientistQA but draws questions from
disjoint entity domains, providing a stricter test of whether a detector learns
a transferable hallucination signal rather than memorizing entity-specific
patterns. We train the detector only on ScientistQA, freeze all preprocessing
and classifier parameters, and evaluate it directly on the new domains. We
highlight three main findings (Table~\ref{tab:detection-auroc}).

\paragraph{I. Perturbation provides strong and transferable signals.}
With Llama, exact perturbation reaches an AUROC of $.910$ on probe-known
ScientistQA, $.949$ on TriviaQA, and $.919$ on DROP, while frozen transfer to
our multi-domain extension reaches $.931$. Similar patterns hold for Mistral and Qwen,
indicating that the perturbation signature is not specific to a single model
family or entity domain. In matched ScientistQA ablations, the full detector
also outperforms an otherwise matched representation without
perturbation-induced features; detailed comparisons are provided in
Appendix~\ref{app:perturb-vs-nonperturb}.

The results are also consistent with the distinguishability account developed above.
On ScientistQA, performance is substantially stronger on the probe-known subset
than on the full benchmark ($.910$ versus $.794$ for the Llama exact detector). When the relevant knowledge is available,
removing a misleading association can shift the model toward the correct key,
producing an informative second-stage response. When such knowledge is weak or
unavailable, the perturbation patterns of correct and erroneous responses become
harder to distinguish.

\paragraph{II. When perturbation is weaker.}
The main exception is GSM8K, where exact perturbation reaches only about
$.740$--$.760$ AUROC across the three models. This is consistent with the
account in Section~\ref{sec:latent-key}: our method is most informative when
the model's error is associated with competition among a small number of localized
input associations. In multi-step mathematical reasoning, however, failure may
arise from distributed computation across many intermediate steps rather than
from one dominant keyword binding. Removing a local span therefore need not
produce the contrastive perturbation response observed in knowledge-known
factual errors. 

\paragraph{III. Reducing the cost of perturbation.}
A limitation of exact two-stage perturbation is its computational cost.
Exhaustively intervening on every candidate span and then repeating the
perturbation after first-stage removal requires many forward passes per
question. We therefore introduce two efficient screening strategies based on
candidate-conditioned attention and input gradients. These signals are used to
rank candidate spans before intervention, so that perturbation is performed
only on the most promising spans. As shown in Table~\ref{tab:perturbation-detection-summary} in Appendix~\ref{app:efficient-perturbation}, both screening
strategies retain most of the performance of exhaustive perturbation across
benchmarks and models. This suggests that the informative perturbation signal
can be captured without enumerating all possible spans. We provide
the detailed algorithms and additional ablations in
Appendix~\ref{app:efficient-perturbation}.

\begin{table*}[t]
\centering
\caption{Hallucination-detection AUROC across method families and benchmarks using Meta-Llama-3.1-8B-Instruct. 
$\dagger$ indicates access to full scientist profiles after answer generation, $\star$ indicates the probe-known subset of ScientistQA, and ``--'' in the Transfer column indicates that frozen-parameter transfer evaluation is not applicable to the method.}
\label{tab:detection-auroc}
\small
\setlength{\tabcolsep}{1.5pt}
\begin{tabular}{llcccccc}
\toprule
Family & Method & ScientistQA$^\star$ & ScientistQA & TriviaQA & GSM8K & DROP & Transfer \\
\midrule
& Ours (Exact) & \textbf{.910} & .794 & \textbf{.949} & .747 & \textbf{.919} & \textbf{.931} \\
Perturbation
& Ours (Attention) & .906 & \textbf{.797} & .946 & .743 & .915 & .922 \\
& Ours (Gradient) & .905 & .790 & .943 & .743 & .915 & .928 \\
\midrule
Representation
& Aiersilan-style & .730 & .646 & .873 & \textbf{.806} & .826 &  .753 \\
& SAPLMA & .671 & .657 & .840 & .719 & .869 & .612 \\
& ICR Probe & .573 & .569 & .726 & .707 & .837 & .563 \\
\midrule
Uncertainty
& SelfCheckGPT & .593 & .580 & .747 & .790 & .765 & -- \\
& SeSE & .560 & .552 & .705 & .632 & .749 & -- \\
& Semantic Entropy & .518 & .528 & .708 & .503 & .752 & -- \\
\midrule
Evidence
& MiniCheck (unilateral) & .717$^\dagger$ & .809$^\dagger$ & .658 & .558 & .604 & -- \\
& MiniCheck (contrastive) & .926$^\dagger$ & .987$^\dagger$ & .823 & .506 & .526 & -- \\
\bottomrule
\end{tabular}

\end{table*}

\section{Hallucination Regimes and Detector Complementarity}
\label{sec:category}

The previous sections focused on one hypothesized failure pattern: the model
possesses the relevant knowledge, but selection of the wrong key is associated
with a competing entity--keyword binding. Keyword perturbation is particularly
effective in this regime. However, Table~\ref{tab:detection-auroc} also shows
that no detector family dominates across all benchmarks. In particular,
representation- and uncertainty-based detection are stronger than perturbation on GSM8K, while
evidence-based verification becomes highly competitive when explicit reference
information is available. We argue that this variation follows naturally from the probabilistic framework
introduced in Section~\ref{sec:latent-key}. Different hallucinations correspond
to failures at different components of the same inference process, and different
detectors observe different components of that process.

\subsection{A Probabilistic Taxonomy of Hallucination Regimes}
\label{sec:probabilistic-taxonomy}

Recall the latent-key decomposition in
Equation~\ref{eq:latent-key-mixture}--\ref{eq:pretraining-key-odds}. From an
empirical perspective, the four regimes differ in which probability is
responsible for the error and therefore in which detector can most directly
expose it.

\begin{enumerate}[leftmargin=*,labelsep=2pt,itemsep=0pt]

\item \textbf{Knowledge-deficit errors.}
Even under the correct key, the answer remains diffuse,
$P(y_r\mid z,x_q,k_r)\approx P(y_w\mid z,x_q,k_r)$. The resulting variation
across outputs or internal trajectories is most naturally captured by
uncertainty- and representation-based detectors; reliable external evidence
can additionally expose a missing fact when one can be supplied.

\item \textbf{Wrong-knowledge errors.}
The model concentrates on an incorrect association or computation,
$P(y_w\mid z,x_q,k_r)\gg P(y_r\mid z,x_q,k_r)$. Since this error can remain
confident and self-consistent, uncertainty is weak; comparison with
external evidence is the most direct detector.

\item \textbf{Context-distraction errors.}
The correct-key answer is reliable,
$P(y_r\mid z,x_q,k_r)\approx 1$, but the context makes a competing key more
probable, $P(k_w\mid z,x_q)>P(k_r\mid z,x_q)$. Perturbation-based detection is
therefore best suited to localized distractions, whereas representation-based
signals can be preferable when the interference is distributed across several
reasoning steps.

\item \textbf{Unstable-inference errors.}
The posterior over competing keys or reasoning trajectories remains diffuse,
$P(k_i\mid z,x_q)\approx P(k_j\mid z,x_q)$ for plausible alternatives, so
small changes in sampling or intermediate computation can change the final
answer. The resulting disagreement and internal variation are most directly
captured by uncertainty- and representation-based detectors.
\end{enumerate}

The mechanisms may overlap; for the analysis below, each response is assigned to one operational regime using
an independent capability probe and repeated generations. The resulting
labels are retrospective analytical annotations: they are not intended to predict 
route examples to detectors, or uniquely identify a latent causal mechanism. Rather, each label
records the dominant failure pattern observable under the benchmark-specific probe.

\subsection{Regime Analysis Across Benchmarks}

Table~\ref{tab:unified-hallucination-types} summarizes the frequencies of the
operational regime labels, which 
provide diagnostic context for the detector rankings in
Table~\ref{tab:detection-auroc}, and here we highlight two representative
cases.

The clearest comparison is within ScientistQA. In the probe-known subset,
context distraction accounts for $56.95\%$ of errors. The probabilistic 
analysis motivates the expectation that perturbation should be particularly effective in this regime because it
directly tests whether a localized competing association controls the answer.
Consistent with this prediction, exact perturbation is
the strongest method that does not access full scientist profiles after
generation. Full ScientistQA instead contains substantially more
knowledge-deficit errors ($42.95\%$) and fewer context-distraction errors
($34.71\%$), so the perturbation signal is expected to weaken. Nevertheless, context distraction still constitutes a large
share of the benchmark, and the perturbation family remains stronger than the
representation- and uncertainty-based methods. Thus, the shift from the known
subset to full benchmark changes the strength, but not the presence, of the
localized dependence captured by perturbation.

GSM8K provides the complementary case. Knowledge deficit and
unstable inference together account for $41.00\%$ of errors, nearly
matching the $47.10\%$ attributed to context distraction. The framework
therefore predicts strong representation and uncertainty signals, as confirmed
by the AUROC of Aiersilan-style detection and
SelfCheckGPT. At the same time, the substantial context-distraction component
leaves a clear perturbation signal ($.747$ AUROC). Because these detector
families target the two major components of the error distribution, their
signals should be complementary. Both combining it with representations and
with uncertainty improves performance ($.821$ AUROC). The fusion results therefore support the
prediction of the taxonomy: a benchmark containing multiple failure
regimes is better covered by combining their corresponding signals than by
single detector family. Complete results are reported in Appendix~\ref{app:category}.
\begin{table*}[t]
\centering
\caption{Distribution of mutually exclusive operational
hallucination labels across benchmarks. Because probe construction and
assignment rules differ by benchmark, percentages are diagnostic and are not
strictly comparable estimates of latent mechanism prevalence (Appendix~\ref{app:category}).}
\label{tab:unified-hallucination-types}
\setlength{\tabcolsep}{4.5pt}
\begin{tabular}{lccccc}
\toprule
Category & Scientist & Scientist$^\star$ & TriviaQA & GSM8K & DROP \\
\midrule
Wrong knowledge & 5.40\% & 0.00\% & 47.60\% & 7.20\% & 8.60\% \\
Knowledge deficit & 42.95\% & 14.57\% & 37.40\% & 21.00\% & 24.00\% \\
Context distract & 34.71\% & 56.95\% & 11.20\% & 47.10\% & 57.00\% \\
Unstable inference & 15.29\% & 26.05\% & 2.40\% & 20.00\% & 9.80\% \\
\bottomrule
\end{tabular}

\end{table*}

\section{Conclusion}
In this work we provide evidence consistent with a latent-key account in which
entity--keyword bindings are associated with hallucinations that arise when
models fail to use accessible knowledge correctly. Answer selection can depend
on these bindings, whose strength varies systematically with pretraining
frequency, and multiple bindings can exhibit competitive and higher-order
interactions. Motivated by this account, we introduced a two-stage
keyword-perturbation detector that
first identifies and removes an influential keyword and then measures how the
remaining evidence reorganizes the model's prediction. The resulting signals
support both hallucination detection and local attribution, while the detector
generalizes across models and benchmarks and transfers to unseen entity
domains. We also find that this advantage is regime-dependent, and hallucination
detection should be matched to the underlying regime.

\section*{AI Use Statement}
In this work, we used generative AI tools to generate synthetic examples for 
our multi-domain dataset, to create and modify scientific figures, 
identify and summarize relevant literature, source and search for information, 
and edit the manuscript to improve readability. We have reviewed all AI-assisted work. 
In particular, AI-generated data were reviewed and filtered before being included 
in the final dataset, and AI-assisted literature search and manuscript edits were 
independently checked by the authors. We have not used generative AI tools to develop 
theoretical models or conceptual frameworks, formulate mathematical claims, 
design research methodology or experiments, implement 
methods, or interpret experimental results; the remaining required-disclosure 
tasks are not applicable to this work.
We take responsibility for the final content of this work, including text, claims, 
or artifacts produced with the aid of generative AI.

\section*{Reproducibility Statement}
To support reproducibility, Appendix~\ref{app:key-odds-proof} states the
assumptions and provides the complete derivation of
Proposition~\ref{prop:key-odds-decomposition}. Appendix~\ref{app:binding-empirical}
specifies the natural-data perturbation operator, matched controls, and
keyword-matching rules, and reports the frequency--dose construction,
training settings, random seeds, and complete results underlying the main
binding analyses; definitions for the interaction analysis are provided in
Appendix~\ref{app:multi}. Appendix~\ref{app:detection} documents the evaluated
checkpoints, detector features, screening procedures, classifier
hyperparameters, data splits, frozen-transfer protocol, and full per-model
metrics, while Appendix~\ref{app:dataset-construction} reports the open-ended
control. Finally, Appendix~\ref{app:category} describes the regime-assignment
procedure, and Table~\ref{tab:empirical-regime-prevalence} reports the complete
benchmark counts. Throughout the appendix, we provide the sample
sizes, random seeds, and evaluation protocols for experiments conducted with
the named public model checkpoints and benchmarks.

\bibliographystyle{unsrtnat}
\bibliography{iclr2027_conference.bib}

\clearpage
\appendix
% !TeX root = iclr2027_conference.tex
% Appendix-ready Related Work section.
% Citation keys are defined in references.bib.

\section{Related Work}
\label{app:introduction}

\paragraph{Hallucination detection from confidence and output variation.}
One major line of work treats hallucination as uncertainty in the model's output distribution.  Token likelihood, perplexity, and verbalized or elicited self-evaluation---including $P(\mathrm{True})$---ask whether the model assigns low confidence to its own answer \citep{kadavath2022mostlyknow}.  Sampling-based methods instead test whether repeated generations agree.  SelfCheckGPT compares a response with stochastic samples without requiring an external knowledge base \citep{manakul2023selfcheckgpt}, while semantic entropy aggregates probability mass over semantically equivalent answers so that paraphrastic variation is not mistaken for epistemic uncertainty \citep{farquhar2024semanticentropy}.  These approaches are powerful for confabulations, but their signal is intrinsically limited for systematic errors: a model can repeatedly produce the same wrong answer with high likelihood.  This distinction is explicit in the scope of semantic entropy, which targets arbitrary, sample-unstable confabulations rather than every kind of factual error.  Our setting centers precisely on the complementary case in which the relevant knowledge is accessible but a statistically salient prompt association consistently dominates the correct constraint.  Hence the relevant question is not only whether the model is uncertain, but which input evidence its wrong preference depends on.

\paragraph{Detection from hidden states and attention.}
White-box detectors exploit information that may be visible internally even when output confidence is misleading.  Early probing work showed that classifiers over hidden activations can distinguish true from false statements \citep{azaria2023internal}.  INSIDE measures consistency among sampled answers in representation space \citep{chen2024inside}; HaloScope learns a truthfulness detector from unlabeled model generations \citep{du2024haloscope}; and ICR Probe models how residual-stream updates evolve across layers rather than reading a single static activation \citep{zhang2025icr}.  Attention-based methods provide a different view.  Lookback Lens predicts contextual hallucination from the ratio of attention paid to the source context versus generated tokens \citep{chuang2024lookback}, whereas LapEigvals uses spectral features of attention maps \citep{binkowski2025spectral}.  These methods establish that error information is encoded in representations and information-flow patterns.  However, an internal state that is predictive of error does not by itself identify the input span responsible for the current decision, and attention weight is not a counterfactual effect.  Our perturbation signature instead measures how the answer preference changes when individual keyword spans are neutralized or replaced, retaining both span identity and effect direction.  The resulting object is simultaneously a detection feature and a local test of evidence dependence.

Recent evaluation work also cautions that strong in-domain detector scores need not imply robust error understanding.  Apparent progress can be inflated by lexical correctness labels or response-length artifacts \citep{janiak2025illusion}, and supervised probes can exploit domain-specific regularities.  We therefore use cross-dataset and frozen-transfer experiments to test whether a perturbation signature captures a reusable dependence pattern rather than merely dataset format.  GSM8K provides a useful boundary condition: chain-of-thought prompting can obscure internal hallucination cues \citep{cheng2025cotobscures}, and arithmetic errors need not be organized around the entity--keyword associations that motivate our method.  We consequently use GSM8K to evaluate the scope of the mechanism rather than as the benchmark from which the mechanism is inferred.

\paragraph{Perturbation-based detection and contextual attribution.}
The nearest methodological family measures a model's response to counterfactual changes in its input.  ContextCite learns a surrogate over subsets of the provided context to attribute a fixed generation to context units \citep{cohenwang2024contextcite}.  GASP removes retrieved chunks and re-scores a fixed answer to detect whether RAG sentences depend on supporting evidence \citep{bouke2026gasp}.  These methods intervene on external context and localize support, but their main target is grounding: whether a generated claim is supported by retrieved or supplied passages.  By contrast, our closed-book and prompt-conditioned setting contains both constraint-relevant and shortcut-inducing cues.  A span may support the correct answer or push the model toward a wrong answer, so the sign of the margin change is essential; absolute sensitivity would conflate these two roles.

Two recent studies are particularly close to our approach.  Shaking to Reveal perturbs prompts with dynamically generated noise and detects hallucinations from changes in intermediate representations \citep{luo2025shaking}.  It establishes that perturbation sensitivity can separate correct and incorrect answers while focusing on representation instability rather than semantically meaningful culprit keywords or their relationship to pretraining frequency.  Sun et al. propose a subsequence-association account in which hallucinatory associations dominate faithful ones and trace causal subsequences by evaluating hallucination probabilities across randomized contexts \citep{sun2025subsequence}.  This account provides a close mechanistic foundation for our view.  Building on these insights, we use targeted, signed, local keyword interventions as a detector across tasks and models, and test frequency-dependent entity--attribute binding with both natural and controlled data.  This connects directional keyword dependence, error detection, and pretraining-frequency imbalance within one empirical framework.

Input perturbation at inference time differs from perturbation-based synthetic-data construction.  HaluGen rewrites system responses to synthesize faithful and hallucinated training examples for a downstream classifier \citep{zhang2024halugen}, whereas our method estimates evidence dependence by perturbing the test prompt.  More generally, deletion, neutralization, and replacement are not interchangeable interventions.  Deletion can alter syntax, discourse structure, or the input distribution, whereas semantic neutralization attempts to preserve form while weakening the targeted association.  We therefore treat agreement across operators as robustness evidence and report operator-specific failures rather than interpreting every change as causal identification.

\paragraph{Pretraining frequency, factual association, and shortcut selection.}
A substantial literature links factual recall to exposure during pretraining.  Accuracy on factual questions rises with the number of relevant pretraining documents \citep{kandpal2023longtail}; entity popularity similarly predicts whether parametric memory is reliable, with retrieval helping most on long-tail facts \citep{mallen2023trust}.  These findings explain which facts are likely to be stored or recalled, but they do not by themselves explain why a model that can answer component probes correctly selects the wrong entity under a richer prompt.  Our focus is therefore not raw rarity alone, but relative association strength among competing entity--keyword bindings.

Recent mechanistic work sharpens this distinction.  Co-occurrence statistics can be learned more readily than abstract factual associations and can fail to transfer to reasoning settings \citep{zhang2024cooccurrence}.  The subsequence-association framework likewise traces dominant prompt--output associations back to the training corpus \citep{sun2025subsequence}.  The latent key--task account of \citet{hu2026reasoningpriors} formalizes a closely related failure mode: pretraining-frequency imbalance increases the posterior weight of a shortcut path relative to a constraint-sensitive path, producing errors even when the relevant knowledge is accessible.  The present work operationalizes that hypothesis at the input-span level.  We test whether frequency changes the strength of entity--keyword binding, whether the same keywords directionally move the wrong-answer margin, and whether the resulting perturbation profile predicts errors.  This closes a gap between corpus-level exposure results, mechanistic accounts of competing associations, and deployable hallucination detection.

\FloatBarrier

\section{Supplementary Material for Section~\ref{sec:binding}}
\label{app:binding}

\subsection{Proof of Proposition~\ref{prop:key-odds-decomposition}}
% !TeX root = iclr2027_conference.tex
\label{app:key-odds-proof}

We derive Proposition~\ref{prop:key-odds-decomposition} as a key-only
specialization of the latent key--task framework of
\citet{hu2026reasoningpriors}.  Their framework models inference through a
posterior over latent key--task pairs,
\begin{equation}
P(y\mid \tilde z)
=
\sum_{k\in\mathcal K,t\in\mathcal T}
P(k,t\mid \tilde z)P(y\mid \tilde z;k,t),
\label{eq:app-full-latent-mixture}
\end{equation}
with a pretraining prior
\begin{equation}
\pi(k,t)
=
\pi^{(k)}(k)\pi^{(t)}(t\mid k).
\label{eq:app-key-task-prior}
\end{equation}
In our setting, we write the full prompt as
$\tilde z=(z,x_q)$, where $z$ denotes the fixed surrounding context and $x_q$
denotes the query content whose association with the competing keys may vary.

Let $(k_r,t_r)$ denote the path associated
with the correct answer $y_r$, and let $(k_w,t_w)$ denote the competing path
associated with $y_w$.  We assume that posterior mass outside these two paths
is negligible.  We further use the output-separation condition that the two
paths induce distinct answers:
\begin{equation}
P(y_r\mid z,x_q;k_w,t_w)\ll 1,
\qquad
P(y_w\mid z,x_q;k_r,t_r)\ll 1.
\label{eq:app-output-separation}
\end{equation}
As in the main text, we focus on the knowledge-available regime in which
\begin{equation}
P(y_r\mid z,x_q;k_r,t_r)\approx 1,
\qquad
P(y_w\mid z,x_q;k_w,t_w)\approx 1.
\label{eq:app-path-confidence}
\end{equation}

Under these conditions, the law of total probability gives
\begin{align}
P(y_r\mid z,x_q)
&\approx
P(k_r,t_r\mid z,x_q)
P(y_r\mid z,x_q;k_r,t_r),
\\
P(y_w\mid z,x_q)
&\approx
P(k_w,t_w\mid z,x_q)
P(y_w\mid z,x_q;k_w,t_w).
\end{align}
Therefore,
\begin{equation}
\frac{P(y_r\mid z,x_q)}
     {P(y_w\mid z,x_q)}
\approx
\frac{P(k_r,t_r\mid z,x_q)}
     {P(k_w,t_w\mid z,x_q)}
\frac{P(y_r\mid z,x_q;k_r,t_r)}
     {P(y_w\mid z,x_q;k_w,t_w)}.
\label{eq:app-output-path-odds}
\end{equation}
By Equation~\ref{eq:app-path-confidence}, the second factor is approximately
one, so
\begin{equation}
\frac{P(y_r\mid z,x_q)}
     {P(y_w\mid z,x_q)}
\approx
\frac{P(k_r,t_r\mid z,x_q)}
     {P(k_w,t_w\mid z,x_q)}.
\label{eq:app-output-path-reduction}
\end{equation}

\begin{equation}
P(k,t\mid z,x_q)
=
P(k\mid z,x_q)P(t\mid k,z,x_q).
\label{eq:app-hierarchical-posterior}
\end{equation}
Our analysis marginalizes task-level variation into the key-conditional
prediction.  Equivalently, within each activated key, the relevant task
posterior is concentrated on its associated task, so that
\begin{equation}
P(t_r\mid k_r,z,x_q)\approx 1,
\qquad
P(t_w\mid k_w,z,x_q)\approx 1.
\end{equation}
Hence,
\begin{equation}
\frac{P(k_r,t_r\mid z,x_q)}
     {P(k_w,t_w\mid z,x_q)}
\approx
\frac{P(k_r\mid z,x_q)}
     {P(k_w\mid z,x_q)}.
\label{eq:app-path-key-reduction}
\end{equation}

It remains to characterize the relative posterior of the two keys.  By Bayes'
rule,
\begin{equation}
P(k\mid z,x_q)
=
\frac{
P(x_q\mid z,k)P(k\mid z)
}{
\sum_{k'}P(x_q\mid z,k')P(k'\mid z)
}.
\label{eq:app-key-bayes}
\end{equation}
Taking the ratio between the two activated keys eliminates the common
normalization:
\begin{equation}
\frac{P(k_r\mid z,x_q)}
     {P(k_w\mid z,x_q)}
=
\frac{P(k_r\mid z)}
     {P(k_w\mid z)}
\frac{P(x_q\mid z,k_r)}
     {P(x_q\mid z,k_w)}.
\label{eq:app-key-posterior-ratio}
\end{equation}

The surrounding context $z$ is fixed across the two competing paths.  We assume that
this fixed context does not itself differentially reweight the two activated
keys.  Their relative prior after conditioning on $z$ therefore follows their
pretraining key frequencies:
\begin{equation}
\frac{P(k_r\mid z)}
     {P(k_w\mid z)}
\propto
\frac{\pi(k_r)}
     {\pi(k_w)},
\label{eq:app-fixed-context-prior}
\end{equation}
where $\pi(k)=P_{\mathrm{pre}}(k)$.

Unlike the fixed context, $x_q$ is precisely the query-dependent evidence whose
association with the competing keys we wish to characterize.  The latent-key
model assumes that this association is inherited from the corresponding
pretraining co-occurrence statistics.  Thus, for the fixed context considered
here,
\begin{equation}
\frac{P(x_q\mid z,k_r)}
     {P(x_q\mid z,k_w)}
\propto
\frac{P_{\mathrm{pre}}(x_q\mid k_r)}
     {P_{\mathrm{pre}}(x_q\mid k_w)}.
\label{eq:app-query-association}
\end{equation}
The proportionality absorbs factors contributed by the fixed surrounding
context that do not vary with $x_q$.

Combining
Equations~\ref{eq:app-output-path-reduction},
\ref{eq:app-path-key-reduction},
\ref{eq:app-key-posterior-ratio},
\ref{eq:app-fixed-context-prior}, and
\ref{eq:app-query-association}, we obtain
\begin{equation}
\frac{P(y_r\mid z,x_q)}
     {P(y_w\mid z,x_q)}
\propto
\frac{\pi(k_r)}
     {\pi(k_w)}
\frac{P_{\mathrm{pre}}(x_q\mid k_r)}
     {P_{\mathrm{pre}}(x_q\mid k_w)}.
\label{eq:app-final-key-odds}
\end{equation}

Finally, using
\begin{equation}
P_{\mathrm{pre}}(k,x_q)
=
P_{\mathrm{pre}}(k)
P_{\mathrm{pre}}(x_q\mid k)
=
\pi(k)P_{\mathrm{pre}}(x_q\mid k),
\end{equation}
Equation~\ref{eq:app-final-key-odds} is equivalently
\begin{equation}
\frac{P(y_r\mid z,x_q)}
     {P(y_w\mid z,x_q)}
\propto
\frac{P_{\mathrm{pre}}(k_r,x_q)}
     {P_{\mathrm{pre}}(k_w,x_q)}
=
\frac{\pi(k_r)}
     {\pi(k_w)}
\frac{P_{\mathrm{pre}}(x_q\mid k_r)}
     {P_{\mathrm{pre}}(x_q\mid k_w)},
\end{equation}
which proves Proposition~\ref{prop:key-odds-decomposition}.
\hfill$\square$
% !TeX root = iclr2027_conference.tex

The proposition is conditional on the two-path, knowledge-available regime
stated in the proof. In particular, it does not claim that pretraining
co-occurrence is the only source of key preference. The fixed-context
proportionalities in Equations~\ref{eq:app-fixed-context-prior} and
\ref{eq:app-query-association} isolate the part of the posterior odds that
changes with the query association; the empirical experiments below test the
corresponding sensitivity, binding, and frequency predictions separately.

\FloatBarrier
\subsection{Empirical Verification}
\label{app:binding-empirical}

\subsubsection{Entity–keyword association reorganization}
\begin{figure}[t]
  \includegraphics[width=0.9\textwidth]{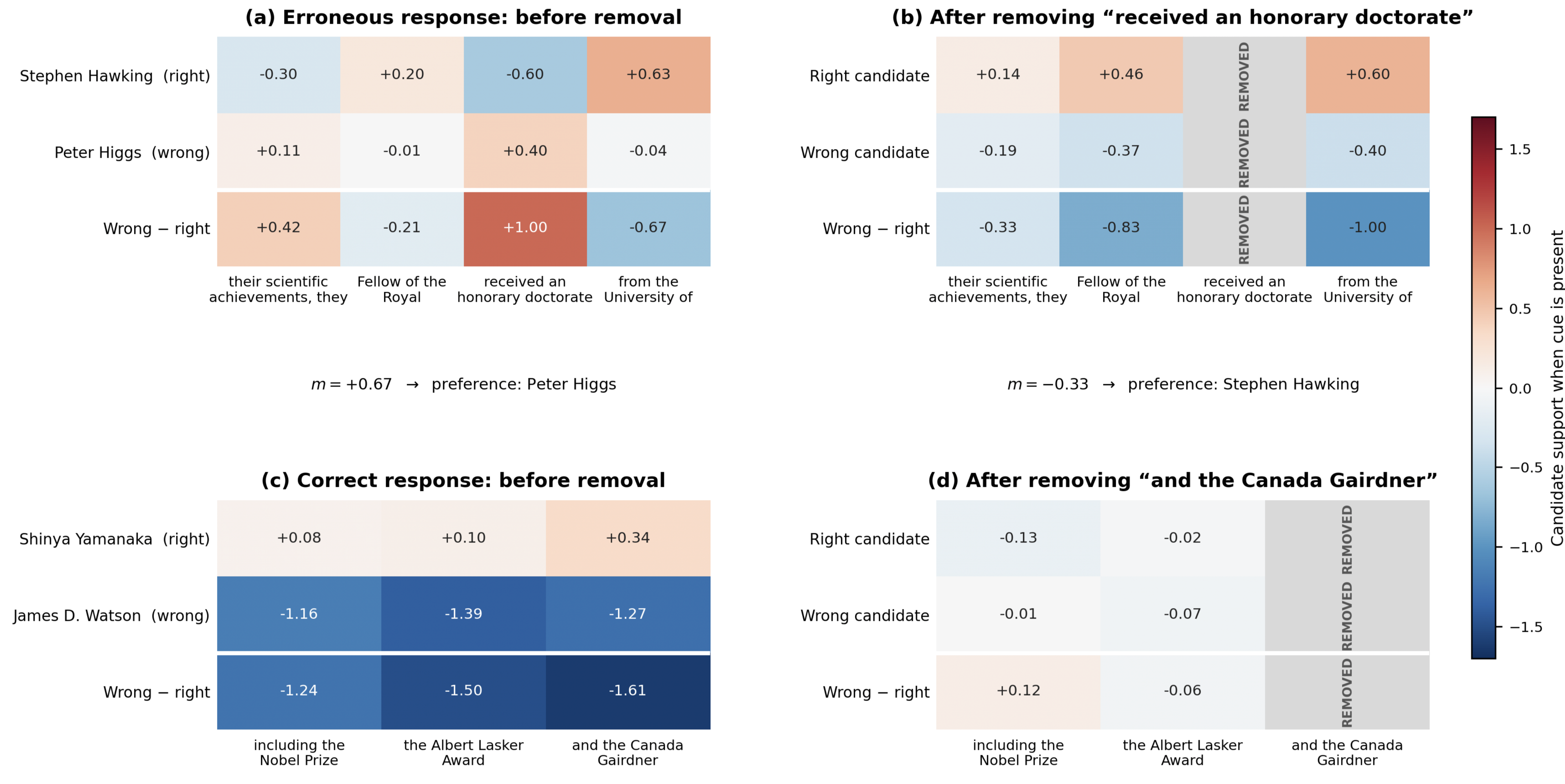}
  \caption{Entity-keyword associations reorganize after perturbation.
Positive values indicate that cue supports candidate.
The final row shows the wrong-minus-right support contrast. In the erroneous example, removing a cue
disproportionately associated with the wrong entity reverses the candidate preference. In the correct example,
removing a cue supporting the right entity reduces a confident correct preference to a near tie. Gray cells
denote the removed cue. Values after removal are recomputed under the perturbed context.}
  \label{fig:association}
\end{figure}

Figure~\ref{fig:association} illustrates how targeted removal
reorganizes the model’s local associations between candidate entities and contextual cues. For each candidate
entity $e$ and cue $j$, we report the cue-support effect, where
a positive value indicates that the presence of the cue increases the candidate’s likelihood. In the erroneous
example (top), the phrase “received an honorary doctorate” disproportionately supports the incorrect candidate,
Peter Higgs, relative to Stephen Hawking, contributing ($+1.00$) to the wrong-minus-right contrast and yielding an
initial margin of ($m=+0.67$). Removing this cue reorganizes the remaining associations and reverses the candidate
preference, producing ($m=-0.33$) in favor of Stephen Hawking. In the correct example (bottom), all three
identified cues strongly favor Shinya Yamanaka, with an initial margin of ($m=-1.63$). Removing the most
influential phrase, “and the Canada Gairdner,” substantially weakens this evidence and moves the margin to
($m=-0.01$), but does not induce a confident preference for the incorrect candidate. These examples provide
qualitative evidence that the perturbation response reflects structured, candidate-specific reliance on
contextual cues rather than a uniform loss of confidence: removing a misleading association can correct an
erroneous preference, whereas removing supporting evidence from a correct response primarily reduces confidence
toward a near tie. This visualization is intended as a mechanistic illustration rather than evidence of
aggregate detection or mitigation performance.

\subsubsection{Local Sensitivity and Entity-Keyword Binding}

\paragraph{Population and model.}
We use the frozen 1,084-item probe-known ScientistQA population. The model is
\texttt{NousResearch/\allowbreak Meta-Llama-3.1-8B-Instruct} in bfloat16. Correctness is
defined by the original exact-name generation and is fixed before any keyword
matching or perturbation is scored. The full population contains 461 erroneous
and 623 correct responses. Candidate scores are length-normalized conditional
log likelihoods, and every effect is oriented using the wrong-minus-right margin
$m(x)=\ell(y_w\mid x)-\ell(y_r\mid x)$.

\paragraph{Semantics-fixed target spans.}
Target construction uses profile/question lexical semantics and does not rank
spans by their measured model effect. We parse both profiles, retain attribute
values that distinguish one profile from the other, and search for conservative
lexical matches in the final natural-language question. The five reported
families are award received, education, field, occupation, and position held.
Matches may contain one to eight words; abstract attributes require full
token-set coverage, while named values require at least 80\% coverage and at
least one informative non-generic token. When several values match the same
token span, the most specific match is retained. Before requiring a control,
this procedure yields 2,409 target records over all supported profile fields,
of which 2,217 belong to the five prespecified families reported here.

\paragraph{Deterministic safe random-span controls.}
For each target, we require a non-target span with exactly the same
\emph{tokenizer-token} length. The eligible pool excludes every target span
plus an eight-token guard band on either side, candidate-name and profile-value
tokens, and relation, negation, or instruction words such as \emph{attended},
\emph{awarded}, \emph{never}, \emph{choose}, and \emph{answer}. Control spans
are selected without replacement, so the same span is never reused within a
question.
We divide the question into four relative-position bins and prefer a control
from the target bin; when none exists, we use the closest available bin.
Ties are ordered by a SHA-256 digest of the protocol version, question ID,
target rank, and control-span ID. Thus the procedure is random with respect to
surface order but fully deterministic and does not inspect model effects.
Every output stores the target and control span IDs, token boundaries, text,
and control-protocol version for audit.

The initial strict same-bin implementation removed many end-loaded questions,
because the guard band could occupy the entire final bin. We therefore use the
bin as a matching preference rather than an exclusion. This final
\texttt{safe-random-v2} rule retains 2,395 target--control records over all
fields and 2,204 records in the five reported families, spanning 1,066
questions (454 erroneous and 612 correct). Three otherwise eligible questions
have no safe span of the exact required token length; they are excluded rather
than weakening the lexical or non-overlap constraints.

Target and control embeddings are independently replaced by the model-wide
mean input embedding while all remaining token embeddings and positions are
held fixed. For any span $s$, the signed neutralization effect is
\begin{equation}
u_s=m(x)-m(x\setminus s).
\label{eq:app-binding-perturbation}
\end{equation}
A positive value means that neutralization lowers the wrong-minus-right margin.
Because this assay tests local dependence rather than direction, the table
reports $|\bar u_s|$: the absolute value of the population-level \emph{signed}
effect, not the mean of per-span absolute effects. We first average spans within
question, then questions within person group, and finally weight person groups
equally. The paired target--control contrast is
$\Delta u=|\bar u_k|-|\bar u_{\mathrm{ctrl}}|$. Confidence intervals use
10,000 person-group bootstrap draws. For $|\bar u_s|$, draws are oriented by
the sign of the observed signed mean before taking quantiles; an interval can
therefore cross zero, as it should under a weak directional effect.

Local sensitivity does not by itself establish an entity-specific binding. We
therefore freeze the natural-data keywords above and probe the reverse
direction: whether conditioning on the keyword's profile owner retrieves that
keyword more often than conditioning on the competing person. For each
keyword, four field-specific prompt templates separately query the owner and
the other person. We draw five continuations per template with temperature
$0.8$, top-$p=0.95$, at most 32 new tokens, and sampling seed 42, giving 20
continuations per person and keyword.

An exact hit occurs when the normalized keyword string appears in the
continuation. The primary owner-aligned statistic is
\begin{equation}
\Delta f_k=
P(k\text{ mentioned}\mid e_{\mathrm{owner}})
-P(k\text{ mentioned}\mid e_{\mathrm{other}}).
\label{eq:app-owner-frequency}
\end{equation}
We additionally computed a token-overlap variant with a small set of
field-specific equivalences, but use exact matching for the paper tables. The
generation outputs were frozen before the safe-control revision; because that
revision changes only the control spans, we reuse those outputs and restrict
scoring to the 2,204 retained target IDs. Exact normalized-string hits are
averaged within item and then within person group, with person groups weighted
equally. The assay therefore covers exactly the same 454 erroneous and 612
correct items as the perturbation table, and its intervals also use 10,000
person-group bootstrap draws.

\begin{table*}[ht]
\centering
\caption{Attribute-keyword perturbation effects, deterministic safe random-span
controls, and owner-aligned keyword retrieval frequencies.
$\lvert\bar u_k\rvert$ and $\lvert\bar u_{\mathrm{ctrl}}\rvert$ are absolute
values of group-averaged signed effects, and
$\Delta u=\lvert\bar u_k\rvert-\lvert\bar u_{\mathrm{ctrl}}\rvert$.
The retrieval frequencies $f_{\mathrm{owner}}$ and $f_{\mathrm{other}}$
condition on the keyword owner and the competing entity, respectively, with
$\Delta f=f_{\mathrm{owner}}-f_{\mathrm{other}}$. Category-specific counts
overlap because a question may contain keywords from multiple attribute
families.}
\label{tab:keyword-binding-controls}
\small
\setlength{\tabcolsep}{5pt}
\begin{tabular}{lrrrrrrr}
\toprule
Attribute & $n$ & $\lvert\bar u_k\rvert$ & $\lvert\bar u_{\mathrm{ctrl}}\rvert$ & $\Delta u$ & $f_{\mathrm{owner}}$ & $f_{\mathrm{other}}$ & $\Delta f$ \\
\midrule
\multicolumn{8}{l}{\textbf{Erroneous responses}} \\
\textbf{Overall} & \textbf{454} & \textbf{.453} & \textbf{.037} & $\mathbf{+.416}$ & \textbf{.414} & \textbf{.078} & \textbf{.337} \\
Award received   & 324 & .427 & .006 & $+.421$ & .492 & .004 & .488 \\
Education        & 144 & .648 & .076 & $+.572$ & .312 & .026 & .287 \\
Field            & 192 & .139 & .048 & $+.091$ & .519 & .329 & .191 \\
Occupation       &  79 & .104 & .066 & $+.038$ & .198 & .117 & .081 \\
Position held    &  63 & .556 & .136 & $+.420$ & .243 & .031 & .212 \\
\midrule
\multicolumn{8}{l}{\textbf{Correct responses}} \\
\textbf{Overall} & \textbf{612} & \textbf{.500} & \textbf{.000} & $\mathbf{+.499}$ & \textbf{.385} & \textbf{.065} & \textbf{.319} \\
Award received   & 374 & .438 & .050 & $+.388$ & .453 & .006 & .447 \\
Education        & 221 & .723 & .105 & $+.618$ & .351 & .013 & .337 \\
Field            & 263 & .225 & .032 & $+.193$ & .509 & .297 & .212 \\
Occupation       & 148 & .215 & .001 & $+.214$ & .231 & .084 & .146 \\
Position held    &  96 & .636 & .001 & $+.635$ & .239 & .026 & .213 \\
\bottomrule
\end{tabular}
\end{table*}

\begin{table}[ht]
\centering
\caption{Keyword perturbation effects and owner-aligned frequency differences
under the safe random-span control. Bracketed values are person-clustered
bootstrap 95\% confidence intervals based on 10,000 draws. Perturbation
intervals are computed for the signed group mean and oriented by its observed
sign; this is why an interval shown beside $\lvert\bar u_k\rvert$ may cross
zero.}
\label{tab:keyword-binding-bootstrap-safe-control}
\small
\setlength{\tabcolsep}{1.5pt}
\begin{tabular}{lccc|ccc}
\toprule
& \multicolumn{3}{c}{\textbf{Erroneous responses}}
& \multicolumn{3}{c}{\textbf{Correct responses}} \\
\cmidrule(lr){2-4}\cmidrule(lr){5-7}
Attribute
& $n$ & $\lvert\bar u_k\rvert$ & $\Delta f$
& $n$ & $\lvert\bar u_k\rvert$ & $\Delta f$ \\
\midrule
\textbf{Overall}
& \textbf{454}
& $\mathbf{0.453}$\,{\scriptsize $[0.388,\,0.524]$}
& $\mathbf{0.337}$\,{\scriptsize $[0.298,\,0.376]$}
& \textbf{612}
& $\mathbf{0.500}$\,{\scriptsize $[0.443,\,0.560]$}
& $\mathbf{0.319}$\,{\scriptsize $[0.289,\,0.351]$} \\
Award received
& 324
& $0.427$\,{\scriptsize $[0.349,\,0.507]$}
& $0.488$\,{\scriptsize $[0.428,\,0.549]$}
& 374
& $0.438$\,{\scriptsize $[0.363,\,0.513]$}
& $0.447$\,{\scriptsize $[0.393,\,0.502]$} \\
Education
& 144
& $0.648$\,{\scriptsize $[0.541,\,0.759]$}
& $0.287$\,{\scriptsize $[0.231,\,0.342]$}
& 221
& $0.723$\,{\scriptsize $[0.637,\,0.817]$}
& $0.337$\,{\scriptsize $[0.292,\,0.383]$} \\
Field
& 192
& $0.139$\,{\scriptsize $[0.065,\,0.217]$}
& $0.191$\,{\scriptsize $[0.136,\,0.246]$}
& 263
& $0.225$\,{\scriptsize $[0.139,\,0.313]$}
& $0.212$\,{\scriptsize $[0.165,\,0.259]$} \\
Occupation
& 79
& $0.104$\,{\scriptsize $[-0.026,\,0.224]$}
& $0.081$\,{\scriptsize $[0.009,\,0.154]$}
& 148
& $0.215$\,{\scriptsize $[0.127,\,0.310]$}
& $0.146$\,{\scriptsize $[0.087,\,0.208]$} \\
Position held
& 63
& $0.556$\,{\scriptsize $[0.380,\,0.748]$}
& $0.212$\,{\scriptsize $[0.149,\,0.279]$}
& 96
& $0.636$\,{\scriptsize $[0.489,\,0.795]$}
& $0.213$\,{\scriptsize $[0.154,\,0.279]$} \\
\bottomrule
\end{tabular}

\end{table}

\paragraph{Results and interpretation.}
The safe controls behave as a genuine placebo: overall
$|\bar u_{\mathrm{ctrl}}|$ is .037 for erroneous responses and rounds to .000
for correct responses, compared with target effects of .453 and .500. Hence
the target--control gaps are .416 and .499. Every family has a positive
point-estimate gap, although the erroneous-response contrasts are smallest for
occupation (.038) and field (.091). Consistent with that weakness, the
orientation-adjusted interval for the erroneous occupation target crosses zero
($[-.026,.224]$); we do not treat that cell alone as conclusive evidence.

The reverse-direction assay is more uniformly entity-specific. Overall
$\Delta f$ is .337 for errors and .319 for correct responses, and every
family-specific 95\% interval is above zero, including the weakest erroneous
occupation cell ($[.009,.154]$). Thus the same phrases that locally affect
candidate preference are preferentially retrieved from their profile owners,
while the safe non-target spans have little aggregate effect. However, target
sensitivity and owner alignment are both comparably strong for correct and
erroneous responses. Binding strength alone is therefore not an error
criterion; the relevant distinction is which candidate a cue supports and how
several signed cue effects combine.

\FloatBarrier
\subsubsection{Pretraining Frequency--Dose Experiment}

The natural-data assays establish association and directional sensitivity but
cannot manipulate the exposure from which a binding is learned. We therefore
construct 50 pairs of fictional people (100 entities) with synthetic names,
institutions, awards, societies, journals, and archives. In an evaluation
prompt, both biographies share an award fact, a society fact, and one
institution cue, either $B$ or $F$. They differ on two crossed facts: the
right candidate has the queried positive fact and satisfies a negative
constraint, whereas the competing candidate has the opposite pattern. The
question is logically identical in the $B$ and $F$ conditions and the selected
institution cue appears in \emph{both} supplied biographies, so the cue is
non-discriminative from the prompt alone.

At each dose $d\in\{0,.25,.5,.75,1\}$, every person contributes $N=40$
training sentences. The weak association occurs $0.05N$ times and the strong
association occurs $(0.05+0.65d)N$ times; remaining sentences contain a neutral
fact. Concretely, the right candidate receives weak $B$ and strong $F$
exposure, while the competing candidate receives strong $B$ and weak $F$
exposure. The assignment is therefore mirrored within every person pair.
Forward forms (person followed by fact) and reverse forms (fact followed by
person) alternate independently within each fact type, using four templates of
each form. The 4,000-sentence corpus is shuffled with the run seed. Dose changes
the relative entity--institution exposure while keeping the people, total
sentences per person, neutral-fact identity, and evaluation prompts fixed.

For every dose and seed, we reload the original checkpoint and fine-tune only
the final four transformer blocks and final normalization layer for two epochs.
We use AdamW with learning rate $10^{-4}$, zero weight decay, batch size 12,
bfloat16, gradient clipping at 1.0, and a maximum training length of 64 tokens;
the causal-language-model loss covers all non-padding tokens. We evaluate
Qwen2.5-3B-Instruct, Llama-3.2-3B-Instruct, and
Ministral-3-3B-Instruct-2512 with seeds 42, 43, and 44.

At evaluation, we score the complete candidate names by length-normalized
conditional log likelihood and average the wrong-minus-right margin over both
profile orders. Let $m_B$ and $m_F$ denote this margin when the shared query cue
is $B$ or $F$. The reported contrast is
$u_\Delta=m_B-m_F$: positive values mean that the cue trained more
often with the competing candidate induces a larger wrong-candidate preference
than the cue trained more often with the right candidate. We fit an ordinary
least-squares slope over the five dose levels separately for each seed.

Table~\ref{tab:binding-dose-three-models_app} gives every seed and the
three-seed mean. At symmetric exposure ($d=0$), the mean contrasts are close to
zero (.028, .002, and .013). At $d=1$, they rise to .232, .064, and .564 for
Qwen, Llama, and Ministral, respectively. All nine model--seed slopes are
positive, with seed-wise Spearman correlations from .6 to 1.0. The effect size
is model dependent---the mean slope is .054 for Llama and .531 for
Ministral---but the exposure direction is stable.

This is a controlled continued-training intervention, not a reconstruction of
the unknown original pretraining corpus. Within that scope, reloading the same
checkpoint at every dose and changing only the mirrored exposure schedule
supports the causal claim that relative entity--keyword frequency can alter
later candidate preference, as predicted for the key-prior term.

\begin{table}[ht]
\centering
\caption{Frequency--dose response across models. Values are candidate-preference differences $u_\Delta$; larger values indicate stronger transfer of asymmetric exposure frequency into candidate preference. Bold values are means across three random seeds; the smaller values to their right report seeds 42, 43, and 44 from top to bottom. The final column reports the seed-wise Spearman correlation $\rho$ between dose and candidate preference in the same order.}
\label{tab:binding-dose-three-models_app}
\small
\setlength{\tabcolsep}{4pt}
\newcommand{\seedcell}[4]{%
  \begin{tabular}[c]{@{}r@{\hspace{2pt}}r@{}}
  & {\scriptsize #2} \\
  \textbf{#1} & {\scriptsize #3} \\
  & {\scriptsize #4}
  \end{tabular}}
\newcommand{\seedrho}[3]{%
  \begin{tabular}[c]{@{}r@{}}
  {\scriptsize #1} \\
  {\scriptsize #2} \\
  {\scriptsize #3}
  \end{tabular}}
\begin{tabular}{lccccccc}
\toprule
Model & $d=0$ & $d=.25$ & $d=.5$ & $d=.75$ & $d=1$ & Slope & $\rho$ \\
\midrule
Qwen2.5-3B
& \seedcell{.028}{.022}{.024}{.039}
& \seedcell{.055}{.028}{.050}{.086}
& \seedcell{.131}{.171}{.073}{.149}
& \seedcell{.200}{.346}{.096}{.156}
& \seedcell{.232}{.286}{.199}{.210}
& \seedcell{.221}{.338}{.159}{.165}
& \seedrho{.9}{1.0}{1.0} \\
\midrule
Llama-3.2-3B
& \seedcell{.002}{-.005}{.000}{.011}
& \seedcell{.013}{.008}{.029}{.003}
& \seedcell{.014}{.009}{.016}{.018}
& \seedcell{.024}{.009}{.029}{.033}
& \seedcell{.064}{.047}{.124}{.021}
& \seedcell{.054}{.042}{.099}{.020}
& \seedrho{1.0}{.9}{.8} \\
\midrule
Ministral-3-3B
& \seedcell{.013}{-.031}{.020}{.050}
& \seedcell{.059}{.074}{.062}{.042}
& \seedcell{.095}{.143}{.121}{.021}
& \seedcell{.285}{.374}{.202}{.279}
& \seedcell{.564}{.589}{.398}{.704}
& \seedcell{.531}{.616}{.359}{.618}
& \seedrho{1.0}{1.0}{.6} \\
\bottomrule
\end{tabular}

\end{table}

\FloatBarrier

\subsection{Interactions among Multiple Keyword Bindings}
\label{app:multi}

\subsubsection{Full-Factorial Interaction Atlas}

We construct candidate cues only from the final ScientistQA question. Exact
occurrences of profile attribute values are grounded to tokenizer spans, and
explicit negation terms (\emph{not}, \emph{never}, \emph{nor},
\emph{without}, and \emph{neither}) are added as separate logic cues. A fixed
semantic priority favors logic terms, locally negated attributes, attributes
owned by only one profile, prespecified field types, and more specific values.
This ordering is fixed before any model effect is measured. Duplicate concepts
and overlapping token spans are removed, and at most six cues are retained in
question order.

For a question with $q$ retained cues, we evaluate all $2^q$ masks (at most 64)
using the same model-wide mean-embedding neutralization and length-normalized
wrong-minus-right margin as above. The frozen 1,084-item generation labels
contain 623 correct and 461 erroneous responses. Eight erroneous items have an
invalid candidate name; another 16 items have fewer than two usable cues (12
correct and four erroneous). The analysed atlas therefore contains 1,060
questions---611 correct and 449 erroneous---and 4,128 cue pairs (2,347 and
1,781, respectively).

Let $u(S)=m(x)-m(x\setminus S)$ be the response when the cue set $S$ is
neutralized, with $u(\varnothing)=0$. For cues $i$ and $j$, the local pair
interaction is
\begin{equation}
I^{\mathrm{local}}_{ij}
=u(\{i,j\})-u(\{i\})-u(\{j\}),
\label{eq:app-local-interaction}
\end{equation}
and its Banzhaf counterpart averages the same second difference over every
background $B$ that contains neither cue,
\begin{equation}
I^{\mathrm{B}}_{ij}
=2^{-(q-2)}\!\!\sum_{B\subseteq[q]\setminus\{i,j\}}
\bigl[u(B\cup\{i,j\})-u(B\cup\{i\})-u(B\cup\{j\})+u(B)\bigr].
\label{eq:app-banzhaf-interaction}
\end{equation}
We also compute the exact Harsanyi dividend for every nonempty set,
\begin{equation}
H(S)=\sum_{T\subseteq S}(-1)^{|S|-|T|}u(T).
\label{eq:app-harsanyi}
\end{equation}
Higher-order mass is the absolute Harsanyi mass at orders three and above,
divided by the absolute mass at all orders two and above.

The prespecified primary threshold is $\tau=.10$; $.05$ and $.20$ serve as
sensitivity checks. A pair is competitive when its two single-cue effects have
opposite signs and both exceed $\tau$ in magnitude. Synergy requires two
positive single-cue effects and both local and Banzhaf interactions above
$\tau$; redundancy requires the corresponding interactions below $-\tau$.
Requiring local and context-averaged interactions to agree avoids labeling an
effect from one arbitrary background as robust non-additivity. All other pairs
are labeled Other. Rates and higher-order mass are computed within question
and then averaged, so every question has equal weight. Confidence intervals
use 10,000 independent question-bootstrap draws within the correct and
erroneous subsets.

At $\tau=.10$, competition occurs in 19.3\% of pairs within erroneous
questions and 13.7\% within correct questions. The 5.5-point difference has a
95\% bootstrap interval of $[2.3,8.7]$ points. Redundancy (9.2\% versus 9.5\%),
synergy (0.6\% versus 0.7\%), and higher-order mass (29.9\% versus 28.5\%) do
not show a comparable error-specific increase: their error-minus-correct
intervals are $[-2.8,2.3]$, $[-.7,.5]$, and $[-1.1,3.9]$ percentage points,
respectively. The competition difference also remains positive at the two
sensitivity thresholds: 26.6\% versus 21.2\% at $\tau=.05$ (difference
$5.3$ points, 95\% CI $[1.6,9.0]$) and 9.9\% versus 6.4\% at $\tau=.20$
(difference $3.6$ points, $[1.2,5.9]$). The robust conclusion is therefore
more specific than generic non-additivity: erroneous responses contain more
strong, oppositely directed cue competition, not simply more interaction mass.

\subsubsection{Frozen Directional Confirmation}

The factorial atlas uses likelihood effects on the full analysis population.
We separately test whether opposite-sign cue pairs cause the predicted changes
in free generation. Before examining confirmation outcomes, we reserve person
groups by SHA-256 parity and use only the parity-one half. Within frozen
generation errors, eligible pairs must have opposite signed single-cue effects
with both magnitudes above .10. A deterministic rule maximizes the weaker
absolute effect, then the stronger effect, with cue IDs as the final tie-break.
This rule selects 92 questions before generation outcomes are inspected. One
question lacks a valid non-overlapping matched control, leaving 91 questions;
we do not relax the control rule to retain it.

For each question we evaluate seven conditions: no intervention, neutralizing
the wrong-supporting cue, neutralizing the right-supporting cue, neutralizing
both, and three width- and question-position-matched random controls. Each
random control uses the exact tokenizer-token width of its target, cannot
overlap any candidate cue or the other control, and is drawn from the same
relative-position quintile when possible. Its RNG seed and token boundaries
are fixed before generation. Each condition uses ten continuations at
temperature .8 and at most 20 new tokens, and all seven conditions share a
per-question sampling seed to make comparisons paired. Exact-name frequencies
are averaged by question; 95\% intervals use 10,000 question-bootstrap draws.
Although Table~\ref{tab:joint-keyword-interventions-app} uses the shorthand
``Remove,'' the implementation uses the same mean-embedding neutralization as
the factorial atlas.

Neutralizing the wrong-supporting cue raises exact right-answer frequency by
14.5 points (95\% CI $[11.0,18.1]$), whereas neutralizing the right-supporting
cue lowers it by 22.5 points ($[-28.4,-16.9]$). Their directional contrast is
37.0 points ($[30.3,43.7]$). Relative to its matched-random control,
neutralizing the wrong-supporting cue yields a 14.4-point additional recovery
($[9.8,19.5]$). These directionally opposed interventions validate the
competition interpretation behaviorally rather than relying only on likelihood
curvature. Joint neutralization does not itself repair the errors:
$P(\mathrm{right})$ is 25.9\%, close to the 27.7\% joint random control and
below the 37.9\% baseline. This further shows why unsigned deletion magnitude
is insufficient: recovery depends on removing the cue that supports the wrong
candidate while preserving the countervailing right-supporting cue.

\begin{table}[ht]
\centering
\caption{Answer probabilities under targeted keyword neutralization and matched
random controls for a frozen confirmation set of 91 baseline errors. Person
groups are reserved by SHA-256 parity before outcomes are inspected; among
opposite-sign pairs with a minimum one-sided effect above 0.10, the pair with
the strongest minimum effect is selected. Probabilities are exact-name
frequencies over ten continuations (temperature 0.8; at most 20 new tokens).
All conditions share a per-question seed. Placebo spans are fixed in advance,
match exact token width and relative question-position quintile when feasible,
and overlap neither candidate cues nor each other.}
\label{tab:joint-keyword-interventions-app}
\small
\setlength{\tabcolsep}{6pt}
\begin{tabular}{lcc}
\toprule
Intervention & $P(\mathrm{right})$ & $P(\mathrm{wrong})$ \\
\midrule
None                                  & 37.9\% & 60.9\% \\
Remove wrong-answer keyword           & 52.4\% & 44.7\% \\
Remove correct-answer keyword         & 15.4\% & 83.5\% \\
Remove both keywords                  & 25.9\% & 70.5\% \\
\midrule
Matched random (wrong-answer side)    & 38.0\% & 60.1\% \\
Matched random (correct-answer side)  & 32.5\% & 65.9\% \\
Matched random (both sides)           & 27.7\% & 70.4\% \\
\bottomrule
\end{tabular}

\end{table}

\FloatBarrier

\section{Supplementary Material for Section~\ref{sec:detection}}
\label{app:detection}

\subsection{Experiment Setup}
Additional analyses below use the two-stage perturbation mechanism defined in
Section~\ref{sec:mechanism}. Unless otherwise stated, all quantities are
computed from a frozen generation: labels indicate whether that generation is
correct, and the two candidate strings are the generated answer and a
benchmark-specific alternative. We use
\texttt{NousResearch/\allowbreak Meta-Llama-3.1-8B-Instruct},\linebreak[4]
\texttt{mistralai/Mistral-7B-Instruct-v0.3},\linebreak[4] and
\texttt{Qwen/Qwen2.5-7B-Instruct}, run in bfloat16 with their native chat
templates. ScientistQA, TriviaQA, DROP, and multidomain generations and labels
are model-specific. Consequently, the probe-known ScientistQA cohorts contain
1,077 Llama, 621 Mistral, and 1,204 Qwen items (335, 267, and 379 person groups,
respectively), and the frozen multidomain target cohorts contain 477, 423, and
350 items. Each TriviaQA cohort contains 1,000 items; its natural correctness
counts are 518, 421, and 538 for Llama, Mistral, and Qwen. DROP is balanced
within model at 500 correct and 500 erroneous generations. The clean GSM8K
stress test instead deliberately fixes the same 942 Llama-generated,
parse-valid responses (471 correct and 471 erroneous) for all three scoring
backbones, so differences there isolate the scoring/representation backbone
rather than generation quality. These population distinctions are important:
the earlier fixed-Llama 1,084-item ScientistQA and 849-item multidomain audits
reported later in this appendix are separate experiments and are not pooled
with the model-specific matrix.

\paragraph{Candidate scoring and perturbation operator.}
For a prompt $x$, generated candidate $y_c$, and alternative $y_a$, we score
each candidate using its length-normalized conditional log likelihood,
\begin{equation}
\ell(y\mid x)=\frac{1}{|y|}\sum_{t=1}^{|y|}
\log p(y_t\mid x,y_{<t}),
\qquad
m_c(x)=\ell(y_c\mid x)-\ell(y_a\mid x).
\label{eq:app-chosen-alternative-margin}
\end{equation}
The candidate span inventory is a deterministic left-to-right partition of
the input context into adjacent non-overlapping two-word spans; a final
one-word span is retained when necessary. To neutralize a span, we replace
the corresponding input-token embeddings by the model-wide mean embedding,
without changing the positions or embeddings of the remaining tokens. If
$x\setminus s$ denotes this intervention, the signed first-stage effect is
\begin{equation}
u_s=m_c(x)-m_c(x\setminus s).
\label{eq:app-stage1-effect}
\end{equation}
Positive $u_s$ means that the span supports the generated candidate relative
to the alternative; negative $u_s$ means that it supports the alternative.

\paragraph{Two-stage collection.}
At the first stage, exact selection evaluates Equation~\ref{eq:app-stage1-effect}
for every candidate span. We retain the five spans with largest $|u_s|$ and
physically delete the strongest one from the text, collapsing excess whitespace
and removing whitespace before punctuation. We then retokenize the resulting
prompt and repeat the full span construction and perturbation procedure. The
five largest second-stage effects describe how candidate preference reorganizes
after the dominant first-stage cue is absent. Candidate likelihoods continue to
be evaluated for the original generated answer and the same alternative, so the
two stages have a common orientation. Items with fewer than five spans are
zero-padded only when the fixed-length feature vector is assembled.

\paragraph{Detection representation.}
Each item is represented by the same fixed 127-dimensional detector input for
all models, benchmarks, and span-selection strategies. The representation has
three components:
\begin{enumerate}[leftmargin=*,labelsep=2pt,itemsep=2pt,topsep=3pt]
    \item \textbf{47 scalar response features.} For both candidates at stage
    one, we include the unperturbed score, the five score changes, their
    versions normalized by the magnitude of the unperturbed score, and summary
    statistics (maximum, minimum, mean absolute effect, standard deviation, and
    positive-effect fraction). For both candidates at stage two, we include the
    unperturbed score and five score changes. Three additional features measure
    the cross-stage change in the chosen score, alternative score, and their
    margin.
    \item \textbf{Four candidate-conditioned hidden-state blocks.} At layer 16,
    we record the unperturbed final-answer-token state for the generated and
    alternative candidates. For each candidate we also compute an
    effect-weighted average hidden-state displacement over the five retained
    first-stage interventions. Each of these four blocks is standardized and
    projected to eight whitened principal components, giving 32 features.
    \item \textbf{Generated-answer state.} We mean-pool the layer-14 states over
    the generated answer tokens, standardize the resulting vector, and retain
    48 whitened principal components.
\end{enumerate}
The scalar block is standardized but not projected. Thus the final dimension is
$47+4\times8+48=127$. All scalers and PCA projections are fitted exclusively on
the current training fold, or exclusively on ScientistQA in frozen transfer.

\paragraph{Efficient span screening.}
The attention and gradient variants alter only which spans receive the costly
counterfactual intervention; downstream feature construction and classifier
training are unchanged. Attention screening averages candidate-conditioned
attention over answer positions, heads, and layers. Its token score is the
attention for the generated candidate plus the absolute generated--alternative
attention difference. We divide the context into 12 contiguous token blocks,
retain the six highest-scoring blocks, and exactly perturb only spans that
overlap those blocks. This screening is recomputed after first-stage deletion.

Gradient screening computes separate derivatives of the generated-candidate
score and alternative-candidate score with respect to a token-wise
neutralization gate; their difference gives the margin derivative. Absolute
span-level deletion gains from the three channels are standardized and summed.
We aggregate spans into punctuation- and newline-delimited sentences, ranking a
sentence by the mean of its three largest span saliencies. Sentences are added
until they cover 75\% of the saliency mass, subject to a cap of 60\% of all
candidate spans, and only spans in the selected sentences are exactly
perturbed. Attention screening requires two candidate-conditioned forward
passes at each stage, while gradient screening requires the corresponding
class-gradient computations.

\paragraph{Dataset construction.}
For ScientistQA and the multidomain benchmark, the alternative is the other
member of the predefined correct/incorrect answer pair. For each TriviaQA
item, the reference answer and a frozen distractor form a candidate pair that
is fixed before model inference and before correctness is evaluated. The model
selects one member as $y_c$, and $y_a$ is always defined mechanically as the
unselected member of this precommitted pair. Consequently, $y_a$ is the
reference answer when the model selects the distractor and the distractor when
it selects the reference answer; this relationship follows from the fixed
candidate pair and does not require the correctness label during candidate
construction or feature extraction. Correctness is computed only afterward
for detector training and evaluation. Because GSM8K and DROP do not provide
alternative candidates, we use
an LLM to construct one for each generated response and freeze it before
feature extraction and detector evaluation. For these two benchmarks, the
benchmark reference or gold answer is used only to determine whether the
original response is correct; it is not used as the alternative candidate in
the paired perturbation scores.
For GSM8K, the evaluated model greedily generates a complete solution ending
in a parseable number, and correctness is exact agreement of that number with
the gold number. We retain a frozen, naturally generated 471/471 subset and
reuse the same responses and LLM-constructed alternatives for all scoring
backbones. For DROP, each model greedily generates a shortest direct answer,
and correctness is normalized exact match to the consensus annotation. A
deterministic seed-42 sample of 500 correct and 500 erroneous generations,
together with their frozen LLM-constructed alternatives, is retained. DROP
groups are passage section IDs; GSM8K groups are source problems.

The multidomain extension contains 1,500 paired-candidate questions, with 500
questions each about athletes, musicians, and buildings. We construct each
item around a predefined correct/incorrect entity pair, following the
ScientistQA format, and retain domain-specific attributes that distinguish the
two entities. We then apply the same independent knowledge probe used for
ScientistQA to determine whether both candidate facts are available to the
evaluated model. As a separate difficulty check, GPT-5.2 answers 464/500
athlete questions (92.8\%), 452/500 questions in the refined musician split
(90.4\%), and 413/500 building questions (82.6\%), for 1,329/1,500 correct
overall (88.6\%). The probe-known rates are matched across domains at 425/500
(85.0\%) each, or 1,275/1,500 (85.0\%) overall. Thus, even for a strong
frontier model, 11.4\% of the questions remain incorrect, indicating that the
extension is not a trivial collection of familiar facts.

\paragraph{Classifier and evaluation protocol.}
We concatenate the scalar and hidden-state blocks and train a class-balanced logistic
regression with $C=0.03$, the \texttt{liblinear} solver, and at most 5,000
iterations. No model- or benchmark-specific detector hyperparameters are tuned
for the matrix reported below. In-domain results use five-fold out-of-fold
prediction repeated with split/PCA seeds 42, 43, and 44. ScientistQA folds are
grouped by person, and GSM8K and DROP folds by source item, using
\texttt{StratifiedGroupKFold}. TriviaQA has one unique source key per item and
therefore uses \texttt{StratifiedKFold}. No group appears in both training and
evaluation within a grouped split. We report the arithmetic mean of the three
per-seed AUROC, AUPRC, and balanced-accuracy values; the fixed operating
threshold for balanced accuracy is 0.5. The positive class is a correct
generation.

For transfer, the scaler, every PCA projection, and the logistic classifier are
fitted on the complete model-specific ScientistQA source cohort and then
frozen. Target-domain labels are used only to compute evaluation metrics.
Target predictions average the three source-side fits corresponding to seeds
42--44; there is no target fitting, calibration, threshold selection, or
hyperparameter tuning. Because randomized PCA depends on the seed, the three
fits can differ even though each uses the full source cohort.

\paragraph{Comparison methods.}
The representation baselines in Table~\ref{tab:detection-auroc} are traced to
three original methods. Aiersilan-style follows the mid-layer hidden-state
linear probe of \citet{aiersilan2026linearly}. SAPLMA is the activation-based accuracy predictor introduced by
\citet{azaria2023internal}, and ICR Probe models layerwise internal-state
dynamics following \citet{zhang2025icr}. The uncertainty baselines are
SelfCheckGPT's sampled-response consistency score
\citep{manakul2023selfcheckgpt}, SeSE's semantic structural entropy
\citep{zhao2026sese}, and Semantic Entropy over meaning-equivalent generations
\citep{farquhar2024semanticentropy}. Finally, both evidence rows use the
published MiniCheck fact-checker \citep{tang2024minicheck}: the unilateral
variant scores the generated answer against the supplied evidence, whereas our
contrastive adaptation subtracts generated-answer support from alternative-
answer support. All baselines operate on the same frozen generations and
correctness labels as our detector.

This evidence access is a substantive difference in the evaluation setting.
As shown in Table~\ref{tab:detection-auroc}, contrastive MiniCheck reaches
.926 AUROC on probe-known ScientistQA and .987 on full ScientistQA, exceeding
the corresponding perturbation results. We therefore do not claim that
perturbation is the optimal detector when reliable ground-truth evidence is
available at inference time: in that setting, direct evidence verification is
the stronger and more natural choice. The value of perturbation is instead in
the evidence-free setting, where no gold profile or reference passage is
assumed and the detector must diagnose the model's dependence on the supplied
query alone. Accordingly, the evidence and perturbation rows should be read as
different information-access regimes rather than as strictly input-matched
competitors.

\paragraph{Perturbation-query accounting.}
Query reduction measures the number of span interventions avoided relative to
enumerating every span at both stages. For attention screening we additionally
charge its four candidate-conditioned screening forwards (two candidates at
each of two stages); the reported reduction therefore does not treat screening
as free. Gradient screening has a different backward-pass cost, so its query
reduction should be interpreted specifically as a reduction in exact span
interventions rather than as an equal-cost wall-clock speedup.

\FloatBarrier
\subsection{Detection Performance}
\label{app:efficient-perturbation}

Table~\ref{tab:perturbation-detection-summary} summarizes the complete
model--benchmark matrix. Exact perturbation is consistently strong on
ScientistQA, TriviaQA, and DROP: exact-selection AUROC spans .822--.910,
.909--.964, and .907--.968, respectively. GSM8K is the clear boundary case:
all three scoring backbones remain near .74--.76 AUROC. This is consistent
with, but does not by itself prove, the hypothesis that many mathematical
errors depend on distributed multi-step computation rather than a localized
entity--keyword association.
The full AUROC, AUPRC, balanced-accuracy, and intervention-count results are
reported in Table~\ref{tab:full-indomain-detection}.

Screening removes a substantial fraction of exact interventions while usually
preserving most of the exhaustive detector's AUROC. For Llama, attention and
gradient screening reduce exact span interventions by roughly 36--47\% and
remain within .007 AUROC of exact selection on every in-domain benchmark. The
tradeoff is more model dependent under domain shift: the Llama transfer gap is
at most .009, whereas Mistral and Qwen lose more under screening. We therefore
view attention and gradient as compute--accuracy operating points rather than
strict replacements for exhaustive selection. Table~\ref{tab:full-transfer-detection}
also shows that transfer difficulty varies by target domain. For example,
Qwen exact selection obtains .911 AUROC on buildings but .724 on musicians.
Because these subsets differ in size, label prevalence, and question
composition, this contrast should be interpreted as heterogeneity across the
frozen target cohorts rather than as a causal effect of entity type.

\begin{table}[ht]
\centering
\caption{AUROC of keyword-perturbation detection across models and benchmarks. In-domain results use repeated five-fold out-of-fold evaluation; Transfer reports aggregate performance from a detector frozen on the corresponding model-specific ScientistQA cohort. $\star$ denotes the probe-known subset of ScientistQA. GSM8K uses the shared clean 942-response cohort described in the text.}
\label{tab:perturbation-detection-summary}
\small
\setlength{\tabcolsep}{6pt}
\begin{tabular}{llccccc}
\toprule
Model & Selection & ScientistQA$^\star$ & TriviaQA & GSM8K & DROP & Transfer \\
\midrule
 & Exact     & .910 & .949 & .7473 & .919 & .931 \\
Llama & Attention & .906 & .946 & .7430 & .915 & .922 \\
 & Gradient  & .905 & .943 & .7426 & .915 & .928 \\
\midrule
 & Exact     & .822 & .964 & .7439 & .968 & .850 \\
Mistral & Attention & .817 & .964 & .7372 & .969 & .803 \\
 & Gradient  & .819 & .966 & .7415 & .965 & .809 \\
\midrule
 & Exact     & .861 & .909 & .7603 & .907 & .848 \\
Qwen & Attention & .835 & .902 & .7575 & .902 & .761 \\
 & Gradient  & .830 & .898 & .7599 & .884 & .811 \\
\bottomrule
\end{tabular}

\end{table}

\begin{table}[ht]
\centering
\caption{Complete in-domain correctness-detection results under repeated five-fold out-of-fold evaluation. Values are means over seeds 42--44. Query reduction is relative to exact two-stage span enumeration; attention includes four screening forwards, while gradient counts avoided exact span interventions.}
\label{tab:full-indomain-detection}
\footnotesize
\setlength{\tabcolsep}{5pt}
\begin{tabular}{lllcccc}
\toprule
Model & Selection & Benchmark & AUROC & AUPRC & Balanced accuracy & Query reduction \\
\midrule
Llama & Exact & ScientistQA & .910 & .930 & .825 & 0.0\% \\
Llama & Exact & TriviaQA & .949 & .956 & .876 & 0.0\% \\
Llama & Exact & GSM8K & .7473 & .7188 & .6872 & 0.0\% \\
Llama & Exact & DROP & .919 & .911 & .836 & 0.0\% \\
Llama & Attention & ScientistQA & .906 & .924 & .825 & 41.4\% \\
Llama & Attention & TriviaQA & .946 & .952 & .877 & 40.6\% \\
Llama & Attention & GSM8K & .7430 & .7191 & .6794 & 35.9\% \\
Llama & Attention & DROP & .915 & .901 & .833 & 46.6\% \\
Llama & Gradient & ScientistQA & .905 & .923 & .817 & 45.4\% \\
Llama & Gradient & TriviaQA & .943 & .950 & .873 & 42.7\% \\
Llama & Gradient & GSM8K & .7426 & .7115 & .6787 & 44.2\% \\
Llama & Gradient & DROP & .915 & .903 & .835 & 43.5\% \\
\midrule
Mistral & Exact & ScientistQA & .822 & .844 & .739 & 0.0\% \\
Mistral & Exact & TriviaQA & .964 & .948 & .896 & 0.0\% \\
Mistral & Exact & GSM8K & .7439 & .7005 & .6929 & 0.0\% \\
Mistral & Exact & DROP & .968 & .959 & .912 & 0.0\% \\
Mistral & Attention & ScientistQA & .817 & .838 & .730 & 40.6\% \\
Mistral & Attention & TriviaQA & .964 & .950 & .897 & 40.0\% \\
Mistral & Attention & GSM8K & .7372 & .6958 & .6886 & 34.5\% \\
Mistral & Attention & DROP & .969 & .961 & .910 & 46.3\% \\
Mistral & Gradient & ScientistQA & .819 & .835 & .729 & 46.6\% \\
Mistral & Gradient & TriviaQA & .966 & .951 & .901 & 41.5\% \\
Mistral & Gradient & GSM8K & .7415 & .7029 & .6791 & 43.9\% \\
Mistral & Gradient & DROP & .965 & .955 & .908 & 42.9\% \\
\midrule
Qwen & Exact & ScientistQA & .861 & .869 & .778 & 0.0\% \\
Qwen & Exact & TriviaQA & .909 & .904 & .829 & 0.0\% \\
Qwen & Exact & GSM8K & .7603 & .7267 & .6996 & 0.0\% \\
Qwen & Exact & DROP & .907 & .909 & .819 & 0.0\% \\
Qwen & Attention & ScientistQA & .835 & .837 & .757 & 40.9\% \\
Qwen & Attention & TriviaQA & .902 & .893 & .825 & 40.1\% \\
Qwen & Attention & GSM8K & .7575 & .7375 & .6957 & 35.5\% \\
Qwen & Attention & DROP & .902 & .902 & .812 & 46.7\% \\
Qwen & Gradient & ScientistQA & .830 & .832 & .750 & 46.6\% \\
Qwen & Gradient & TriviaQA & .898 & .894 & .820 & 43.6\% \\
Qwen & Gradient & GSM8K & .7599 & .7213 & .6925 & 45.2\% \\
Qwen & Gradient & DROP & .884 & .877 & .794 & 47.0\% \\
\bottomrule
\end{tabular}
\end{table}

\begin{table}[ht]
\centering
\caption{Complete model-specific multidomain transfer results for detectors frozen on ScientistQA. Source/target sizes for Llama, Mistral, and Qwen are 1,077/477, 621/423, and 1,204/350. Target labels are never used for fitting or calibration.}
\label{tab:full-transfer-detection}
\footnotesize
\setlength{\tabcolsep}{5pt}
\begin{tabular}{lllcccc}
\toprule
Model & Selection & Target & AUROC & AUPRC & Balanced accuracy & Query reduction \\
\midrule
Llama & Exact & All & .931 & .972 & .791 & 0.0\% \\
Llama & Exact & Athlete & .955 & .988 & .789 & 0.0\% \\
Llama & Exact & Building & .945 & .981 & .852 & 0.0\% \\
Llama & Exact & Musician & .897 & .935 & .728 & 0.0\% \\
Llama & Attention & All & .922 & .967 & .785 & 31.7\% \\
Llama & Attention & Athlete & .949 & .985 & .785 & 31.7\% \\
Llama & Attention & Building & .934 & .974 & .849 & 31.7\% \\
Llama & Attention & Musician & .889 & .930 & .723 & 31.7\% \\
Llama & Gradient & All & .928 & .970 & .767 & 45.4\% \\
Llama & Gradient & Athlete & .947 & .985 & .757 & 45.4\% \\
Llama & Gradient & Building & .940 & .976 & .815 & 45.4\% \\
Llama & Gradient & Musician & .896 & .936 & .723 & 45.4\% \\
\midrule
Mistral & Exact & All & .850 & .942 & .679 & 0.0\% \\
Mistral & Exact & Athlete & .872 & .963 & .764 & 0.0\% \\
Mistral & Exact & Building & .856 & .947 & .672 & 0.0\% \\
Mistral & Exact & Musician & .860 & .928 & .635 & 0.0\% \\
Mistral & Attention & All & .803 & .920 & .659 & 30.0\% \\
Mistral & Attention & Athlete & .829 & .947 & .712 & 30.0\% \\
Mistral & Attention & Building & .793 & .923 & .644 & 30.0\% \\
Mistral & Attention & Musician & .832 & .909 & .647 & 30.0\% \\
Mistral & Gradient & All & .809 & .919 & .691 & 43.8\% \\
Mistral & Gradient & Athlete & .806 & .933 & .711 & 43.8\% \\
Mistral & Gradient & Building & .842 & .945 & .696 & 43.8\% \\
Mistral & Gradient & Musician & .826 & .890 & .688 & 43.8\% \\
\midrule
Qwen & Exact & All & .848 & .929 & .671 & 0.0\% \\
Qwen & Exact & Athlete & .837 & .924 & .675 & 0.0\% \\
Qwen & Exact & Building & .911 & .971 & .738 & 0.0\% \\
Qwen & Exact & Musician & .724 & .794 & .589 & 0.0\% \\
Qwen & Attention & All & .761 & .876 & .614 & 30.8\% \\
Qwen & Attention & Athlete & .754 & .867 & .605 & 30.8\% \\
Qwen & Attention & Building & .815 & .933 & .660 & 30.8\% \\
Qwen & Attention & Musician & .658 & .720 & .564 & 30.8\% \\
Qwen & Gradient & All & .811 & .906 & .652 & 44.5\% \\
Qwen & Gradient & Athlete & .791 & .896 & .619 & 44.5\% \\
Qwen & Gradient & Building & .872 & .954 & .743 & 44.5\% \\
Qwen & Gradient & Musician & .700 & .766 & .593 & 44.5\% \\
\bottomrule
\end{tabular}
\end{table}

Transfer AUROC can exceed source-domain out-of-fold AUROC, but the two values
are not estimates on the same population: the target cohort is conditioned on
the model knowing both candidate entities and has different label prevalence
and domain composition. We therefore use transfer only as evidence that the
frozen decision rule retains ranking signal outside ScientistQA, not as
evidence that the target task is intrinsically easier.

\FloatBarrier
\subsection{Open-Ended Generation Version}
\label{app:dataset-construction}

The main experiments use the generated answer and a benchmark-specific
alternative to form a contrastive margin. We additionally test a stricter
open-ended setting in which the alternative answer is never loaded by the
perturbation or representation pipeline. This control asks whether the method
can detect an erroneous free-form generation from the behavior of that
generation alone, rather than from access to a known competing answer.

\paragraph{Pred-only perturbation.}
Let $y_c$ be the model's original free-form generation. For every span $s$, the
open-ended perturbation effect is
\begin{equation}
u_s^{\mathrm{open}}
=
\ell(y_c\mid x)-\ell(y_c\mid x\setminus s).
\label{eq:open-ended-effect}
\end{equation}
Span ranking, first-stage deletion, and the second-stage rescan use only
$|u_s^{\mathrm{open}}|$. The feature vector likewise contains only the
generated-answer score trajectory, its layer-16 unperturbed state and
effect-weighted displacement, and its layer-14 answer state. In particular,
alternative-answer scores, alternative-answer hidden states, contrastive
margins, and contrastive span ranking are all excluded. The same generated
answer is teacher-forced before and after perturbation; setting the unused
second candidate in the shared implementation equal to $y_c$ is only a
container requirement, and its returned score is discarded.

The resulting detector has 87 inputs: 23 generated-answer scalar features, an
eight-component unperturbed layer-16 block, an eight-component
effect-weighted layer-16 displacement block, and 48 mean-pooled layer-14
components. The benchmark rows, labels, two-stage deletion operator, folds,
seeds 42--44, fold-local preprocessing, and balanced logistic regression with
$C=.03$ remain fixed across model families.

Table~\ref{tab:open-ended-p-grouped} reports the strict pred-only detector for
all three model families together with the corresponding dual-candidate exact
reference. Removing the alternative reduces Llama AUROC by .036 on
ScientistQA, .069 on TriviaQA, and .046 on DROP, but not on GSM8K, where
pred-only is .011 higher. The same exception appears for Mistral: pred-only is
.047 higher on GSM8K while remaining lower on the other three benchmarks.
Thus the competing answer is useful in most settings but is not the sole
source of the detection signal. Pred-only detection also remains informative
for Mistral and Qwen, with macro-average AUROCs of .856 and .781,
respectively. The GSM8K results caution against treating the dual-candidate
construction as uniformly easier: two long solution strings can add
contrastive variation without adding a localized correctness cue.

\begin{table}[h]
\centering
\caption{Open-ended, pred-only perturbation detection and its same-row
dual-candidate exact reference. Exact, pred-only, and $\Delta$ report AUROC;
AUPRC and balanced accuracy are computed for the pred-only detector with
correctness as the positive class. Values are means over seeds 42--44 under
repeated five-fold OOF evaluation, grouped by person/source item except for
unique-key TriviaQA. These are the fixed legacy Llama cohorts; in particular,
the 1,084-row ScientistQA exact reference (.902) differs from the model-specific
1,077-row primary result (.910). The Mistral and Qwen exact references are the
model-specific results reported in Table~\ref{tab:full-indomain-detection}.}
\label{tab:open-ended-p-grouped}
\footnotesize
\setlength{\tabcolsep}{4pt}
\begin{tabular}{llrrrrrr}
\toprule
Model & Benchmark & $n$ & Exact & Pred-only & $\Delta$ & AUPRC & Bal.\ acc. \\
\midrule
Llama & ScientistQA$^\star$ & 1,084 & .902 & .866 & $-.036$ & .893 & .781 \\
      & TriviaQA              & 1,000 & .965 & .896 & $-.069$ & .895 & .813 \\
      & GSM8K                 &   942 & .743 & .754 & $+.011$ & .725 & .692 \\
      & DROP                  & 1,000 & .919 & .873 & $-.046$ & .834 & .800 \\
\midrule
Mistral & ScientistQA$^\star$ &   621 & .822 & .736 & $-.086$ & .778 & .672 \\
        & TriviaQA              & 1,000 & .964 & .943 & $-.021$ & .911 & .878 \\
        & GSM8K                 &   942 & .744 & .791 & $+.047$ & .611 & .727 \\
        & DROP                  & 1,000 & .968 & .956 & $-.012$ & .942 & .901 \\
\midrule
Qwen    & ScientistQA$^\star$ & 1,204 & .861 & .758 & $-.103$ & .784 & .696 \\
        & TriviaQA              & 1,000 & .909 & .885 & $-.024$ & .884 & .795 \\
        & GSM8K                 &   942 & .760 & .754 & $-.006$ & .891 & .695 \\
        & DROP                  & 1,000 & .907 & .728 & $-.179$ & .698 & .671 \\
\bottomrule
\end{tabular}
\end{table}

\subsection{Ordinary 80/20 Splits and Entity Leakage}
\label{app:random-split}

We also audit the sensitivity of the ScientistQA result to the splitting unit.
An ordinary stratified 80/20 split assigns individual questions independently.
Because ScientistQA contains several questions about the same person, this
places questions sharing an entity---and therefore much of their lexical,
factual, and representation structure---on both sides of the split. We regard
this as \emph{entity-level data leakage}: preprocessing is fitted on the
training partition and no exact row is duplicated, but the test entities are no
longer independent of the training entities. A representation probe can exploit
person-specific regularities rather than learn a domain-general correctness
signal.

Table~\ref{tab:split-protocol-sensitivity} quantifies the effect under matched
Llama detector configurations. The grouped values average seeds 42--44 over
five-fold OOF; the ordinary 80/20 values average six independent holdouts
(seeds 42--47). Aiersilan-style rises from .730 to .904 AUROC on probe-known
ScientistQA and from .646 to .810 on full ScientistQA when person grouping is
removed, gains of 17.4 and 16.4 points. Exact perturbation changes by 0.6 and
2.5 points, respectively. The larger Aiersilan-style increase is compatible
with entity overlap inflating the ordinary-split estimate, although the change
from OOF to a single holdout is a second protocol difference. Consequently, we
use person-group OOF for the main ScientistQA claims and keep ordinary 80/20
results only as explicitly labeled supplementary analyses.

 \begin{table*}[t]
  \centering
  \caption{ScientistQA AUROC sensitivity to person-group versus example-random
  splitting. Exact perturbation uses ordinary stratified 80/20 splits; the
  Aiersilan-style MLP uses its official 70/10/20 train/validation/test split.
  Within each row, the model, representation, layer, preprocessing, probe, and
  test fraction are held fixed.}
  \label{tab:split-protocol-sensitivity}
  \small
  \setlength{\tabcolsep}{4pt}
  \begin{tabular}{llcc}
  \toprule
  Model & Subset and method & Person-group OOF & Example-random holdout \\
  \midrule
  Llama
    & Probe-known, exact perturbation    & .908 & .917 \\
    & Probe-known, Aiersilan-style MLP   & .765 & .904 \\
    & Full, exact perturbation           & .794 & .819 \\
    & Full, Aiersilan-style MLP          & .626 & .808 \\
  \midrule
  Mistral
    & Probe-known, exact perturbation    & .824 & .846 \\
    & Probe-known, Aiersilan-style MLP   & .681 & .867 \\
    & Full, exact perturbation           & .739 & .789 \\
    & Full, Aiersilan-style MLP          & .585 & .788 \\
  \midrule
  Qwen
    & Probe-known, exact perturbation    & .861 & .887 \\
    & Probe-known, Aiersilan-style MLP   & .713 & .863 \\
    & Full, exact perturbation           & .720 & .768 \\
    & Full, Aiersilan-style MLP          & .565 & .783 \\
  \bottomrule
  \end{tabular}
  \end{table*}

This split-only audit was not completed for attention, gradient, Mistral, or
Qwen. We therefore do not extrapolate those cells from the primary grouped
matrix. Their main person-disjoint results remain available in
Table~\ref{tab:full-indomain-detection}.

\FloatBarrier

\subsection{Frozen Transfer to New Entity Domains}
\label{app:multidomain-transfer}

The primary model-specific transfer experiment is already reported in
Table~\ref{tab:full-transfer-detection}. We additionally retain an earlier
fixed-Llama audit because it places several representation baselines on one
common target. In this audit every detector is fitted on ScientistQA, frozen,
and evaluated without target fitting or tuning on 849 \emph{knows-both}
questions: 316 athlete, 264 musician, and 269 building items. The target
contains 686 correct and 163 erroneous answers and has no source--target item
key overlap. The three source-side fits use seeds 42--44 and their target
probabilities are averaged.

This fixed audit uses either the legacy 1,084-item probe-known source or the
2,894-item full ScientistQA source. It is not row-compatible with the
model-specific transfer cohorts in Table~\ref{tab:full-transfer-detection};
although exact perturbation obtains .931 aggregate AUROC in both audits, the
identical value refers to different source and target populations. All three
perturbation variants outperform the representation baselines on every target
slice in Table~\ref{tab:scientist-multidomain-transfer}. From the probe-known
source, exact, attention, and gradient perturbation obtain aggregate AUROCs of
.931, .922, and .928, respectively, compared with .677 for Aiersilan-style,
.606 for SAPLMA, and .471 for ICR Probe. The advantage persists when training
on full ScientistQA: the three perturbation variants obtain .923, .909, and
.903, whereas the strongest representation baseline, SAPLMA, reaches .610.
These results support transfer of the perturbation trajectory on this fixed
target, but they do not identify person overlap as the sole cause of the
baseline gap.

  \begin{table}[h]
  \centering
  \caption{AUROC for strict frozen transfer from fixed Llama ScientistQA sources to a common 849-item multidomain target. Every transform and classifier is fitted exclusively on the indicated source population; target labels are used only to compute metrics. Exact perturbation uses the 127-dimensional representation. Aiersilan-style uses the generated-answer layer-14 last-token state; SAPLMA and ICR use their published MLP training protocols.}
  \label{tab:scientist-multidomain-transfer}
  \footnotesize
  \setlength{\tabcolsep}{4pt}
  \begin{tabular}{llcccc}
  \toprule
  Source population & Method & All & Athlete & Musician & Building \\
  \midrule
  Probe-known
  & Exact perturbation & \textbf{.931} & \textbf{.955} & \textbf{.897} & \textbf{.945} \\
  & Perturbation (attention) & .922 & .949 & .889 & .934 \\
  & Perturbation (gradient) & .928 & .947 & .896 & .940 \\
  & Aiersilan-style & .677 & .727 & .667 & .682 \\
  & SAPLMA & .606 & .645 & .602 & .640 \\
  & ICR Probe & .471 & .485 & .402 & .530 \\
  \midrule
  Full ScientistQA
  & Exact perturbation & \textbf{.923} & \textbf{.946} & \textbf{.881} & \textbf{.942} \\
  & Perturbation (attention) & .909 & .942 & .880 & .913 \\
  & Perturbation (gradient) & .903 & .924 & .874 & .916 \\
  & Aiersilan-style & .598 & .637 & .583 & .617 \\
  & SAPLMA & .610 & .675 & .583 & .683 \\
  & ICR Probe & .444 & .394 & .483 & .443 \\
  \bottomrule
  \end{tabular}
  \end{table}

\begin{table}[ht]
\centering
\caption{AUROC of Aiersilan hidden-state features and their fusion with
perturbation hidden states (excluding non-perturbed hidden states) for
in-domain ScientistQA evaluation and frozen multidomain transfer. Fusion gain
is measured in AUROC points relative to the Aiersilan hidden-state baseline.}
\label{tab:aiersilan-perturb-hidden-only-fusion}
\small
\setlength{\tabcolsep}{4pt}
\begin{tabular}{lrrr}
\toprule
Setting & Aiersilan hidden & Fusion & Gain over Aiersilan \\
\midrule
\multicolumn{4}{l}{\textit{Panel A: In-domain and transfer results}} \\
ScientistQA in-domain & .7286 & .8616 & +13.30 pts \\
Multidomain transfer  & .7577 & .9318 & +17.41 pts \\
\midrule
\multicolumn{4}{l}{\textit{Panel B: Multidomain transfer by target domain}} \\
Athlete  & .7340 & .9526 & +21.86 pts \\
Musician & .7477 & .8958 & +14.81 pts \\
Building & .7759 & .9481 & +17.22 pts \\
Overall  & .7577 & .9318 & +17.41 pts \\
\bottomrule
\end{tabular}
\end{table}

The frozen transfer AUROC is higher than the corresponding in-domain
ScientistQA AUROC despite the absence of target-domain fitting. We conjecture
that this difference primarily reflects target composition rather than a
benefit from distribution shift. The transfer set contains many prominent
athletes, musicians, and buildings that are likely to be well represented in
pretraining, and it is restricted to questions for which the model knows both
candidate entities. The model is therefore more likely to possess the relevant
facts even when its original answer is wrong. In this knowledge-available
regime, removing a misleading keyword can expose a stable recovery toward the
correct candidate, producing a cleaner perturbation signal than on
ScientistQA, which includes more obscure entities and knowledge-deficit cases.
This explanation remains a hypothesis because we do not directly measure the
pretraining exposure of individual target entities.

Attention and gradient transfer for all three backbones belong to the primary
model-specific experiment and are given in
Table~\ref{tab:full-transfer-detection}. We do not copy those values into the
849-item table because doing so would create a false same-row comparison.

\FloatBarrier

\subsection{Runtime and Resource Efficiency}
\label{app:runtime-efficiency}

We benchmark the three span-selection strategies using
Llama-3.1-8B-Instruct in bfloat16 on a single NVIDIA A100 40GB GPU with batch
size 24. Two independent items are run as warm-up and excluded from the timing
summary. All methods otherwise use the same examples and implementation, so
the latency differences isolate the selection and intervention workloads.

\begin{table}[ht]
\centering
\caption{Wall-clock and resource measurements for exact enumeration,
attention screening, and gradient screening. Latency is measured per item;
incremental peak memory is reported in GiB. Speedup and time reduction are
relative to Exact.}
\label{tab:runtime-resource-efficiency}
\small
\setlength{\tabcolsep}{8pt}
\begin{tabular}{lrrr}
\toprule
Metric & Exact & Attention & Gradient \\
\midrule
Query reduction & 0\% & 46.8\% & 48.1\% \\
Seconds/item (mean $\pm$ SD) & $2.653 \pm .764$ & $1.675 \pm .434$ & $1.822 \pm .428$ \\
P95 seconds/item & 4.035 & 2.458 & 2.592 \\
Speedup vs. Exact & $1.00\times$ & $1.58\times$ & $1.46\times$ \\
Time reduction & 0\% & 36.9\% & 31.3\% \\
Forward calls/item & 10.28 & 10.70 & 10.55 \\
Backward calls/item & 0 & 0 & 2 \\
Incremental peak memory & 0.963 & 0.840 & 1.756 \\
\bottomrule
\end{tabular}
\end{table}

\begin{table}[ht]
\centering
\caption{Effective batched workload for the runtime benchmark. Teacher-forced
rows include the candidate-scoring rows in addition to perturbation rows. Time
saved is the paired per-item difference relative to Exact.}
\label{tab:runtime-effective-workload}
\small
\setlength{\tabcolsep}{7pt}
\begin{tabular}{lrrr}
\toprule
Method & Perturbation rows/item & Teacher-forced rows/item & Time saved/item \\
\midrule
Exact     & 74.16 & 76.16 & -- \\
Attention & 39.48 & 41.48 & 0.979 s \\
Gradient  & 38.47 & 40.47 & 0.832 s \\
\bottomrule
\end{tabular}
\end{table}

The query reductions translate into substantial, but not one-to-one,
wall-clock gains. Attention screening removes 46.8\% of perturbation queries
and reduces mean latency by 36.9\%, corresponding to a $1.58\times$ speedup.
Its paired saving is 0.979 seconds per item (95\% CI $[0.958,0.999]$), and it
removes an average of 34.68 intervention rows. Gradient screening removes
slightly more queries (48.1\%, or 35.70 intervention rows) but reduces latency
by 31.3\%, corresponding to a $1.46\times$ speedup. Its paired saving is 0.832
seconds per item (95\% CI $[0.811,0.852]$). The two backward passes required
for gradient scoring explain why its larger query reduction produces a smaller
speedup than attention screening and why its incremental peak memory is
higher.

Attention screening also reduces incremental peak memory from 0.963 to 0.840
GiB, approximately 12.7\% below Exact. Forward-call counts do not decrease
because they count top-level batched invocations rather than the number of rows
processed within each invocation. The relevant workload reduction instead
appears in the perturbation and teacher-forced row counts in
Table~\ref{tab:runtime-effective-workload}. Reporting these row counts together
with wall-clock latency therefore gives a more faithful account of realized
efficiency than top-level forward calls alone.

\FloatBarrier

\FloatBarrier
% !TeX root = iclr2027_conference.tex
\section{Supplementary Materials for Ablation Studies}
\label{app:ablations}

This appendix isolates four design choices in the detector: the second
perturbation stage, the use of perturbation responses beyond the unperturbed
margin, the contribution of perturbation-induced hidden states, and targeted
keyword selection relative to random spans. We additionally evaluate direct
transfer between two distinct benchmarks. Unless stated otherwise, all
comparisons use the same benchmark- and model-specific cohorts, labels,
grouping rules, and out-of-fold splits as the corresponding full detector.

\subsection{Stage-1 versus Two-Stage Detection}
\label{app:stage-ablation}

\begin{table}[H]
\centering
\caption{Stage-1 and full two-stage detection under matched model-specific
evaluation protocols. Stage 1 retains the initial-scan features, whereas the
full detector adds the second-stage and cross-stage features. Values are means
over seeds 42--44. Query reduction is measured relative to exact enumeration
within the corresponding stage; dashes in the exact rows denote the reference
method.}
\label{tab:stage-ablation}
\small
\setlength{\tabcolsep}{3.5pt}
\begin{tabular}{lllrrrrrr}
\toprule
& & & \multicolumn{3}{c}{Stage 1} & \multicolumn{3}{c}{Two-stage} \\
\cmidrule(lr){4-6}\cmidrule(lr){7-9}
Model & Benchmark & Selection
& AUROC & AUPRC & Query red.
& AUROC & AUPRC & Query red. \\
\midrule
Llama & ScientistQA & Exact     & .886 & .878 & -- & .910 & .930 & -- \\
      &             & Attention & .884 & .869 & 37.1\% & .906 & .924 & 41.4\% \\
      &             & Gradient  & .879 & .865 & 47.2\% & .904 & .923 & 45.4\% \\
\midrule
Mistral & ScientistQA & Exact     & .801 & .826 & -- & .822 & .844 & -- \\
        &             & Attention & .806 & .833 & 37.4\% & .817 & .838 & 40.6\% \\
        &             & Gradient  & .796 & .822 & 47.2\% & .819 & .835 & 46.6\% \\
\midrule
Qwen & ScientistQA & Exact     & .843 & .849 & -- & .861 & .869 & -- \\
     &             & Attention & .829 & .831 & 37.9\% & .835 & .837 & 40.9\% \\
     &             & Gradient  & .825 & .827 & 47.8\% & .830 & .832 & 46.6\% \\
\bottomrule
\end{tabular}
\end{table}

The consistent gains from the second stage raise a complementary question:
why is the stage-1-only detector already effective? Perturbing an influential
keyword can substantially change the model's generation, particularly its
answer margin and related response features. Under our latent-key account,
neutralizing a keyword that supports a correct answer typically produces a
smaller and less systematic response than neutralizing a misleading keyword
that drives an erroneous answer. In the latter case, suppressing the misleading
association can redirect the model toward the correct keyword and may even
reverse its answer. The first-stage features therefore already contain a strong
discriminative signal, especially for examples in which the model is readily
misled by a dominant association. The second stage adds complementary
information by revealing structure that the initial intervention cannot expose:
after the dominant keyword is removed, the detector can identify the
next-most-influential association and measure the resulting reorganization of
the model's preference. These additional responses account for the consistent
improvement of the full two-stage detector.

\iffalse
\begin{table}[ht]
\centering
\caption{Macro-average AUROC across the three backbones for the Stage-1 and
full two-stage detectors on TriviaQA and GSM8K. The last column is Full minus
Stage 1.}
\label{tab:stage-ablation-trivia-gsm8k}
\small
\setlength{\tabcolsep}{7pt}
\begin{tabular}{llrrr}
\toprule
Benchmark & Selection & Stage 1 & Full & Difference \\
\midrule
TriviaQA & Exact     & .9422 & .9408 & $-.0015$ \\
         & Attention & .9386 & .9371 & $-.0016$ \\
         & Gradient  & .9378 & .9359 & $-.0020$ \\
\midrule
GSM8K & Exact     & .7495 & .7505 & $+.0011$ \\
      & Attention & .7478 & .7459 & $-.0020$ \\
      & Gradient  & .7487 & .7480 & $-.0007$ \\
\bottomrule
\end{tabular}
\end{table}

On ScientistQA, the second stage improves AUROC in all nine matched
comparisons: .022--.025 for Llama, .011--.023 for Mistral, and .005--.008 for
Qwen; AUPRC also increases throughout. This pattern does not extend to the two
additional benchmarks. TriviaQA slightly favors Stage 1 under all three
selection methods, while the GSM8K differences remain within approximately
$\pm .002$ and change direction across models and selection methods. Stage 1
also reduces the query count by approximately 49\% relative to its matched
two-stage detector. Thus, the second stage defines the mechanistically useful
recovery trajectory after removing the strongest first-stage cue, but does not
provide a stable improvement in in-domain ranking performance across
benchmarks.
\fi

\subsection{Margin-Only Control}
\label{app:margin-only-ablation}

The margin-only control retains the unperturbed answer margin but removes all
perturbation-response features. It is evaluated on exactly the same rows,
labels, groups, and fold assignments as the full detector, so the comparison
isolates the additional information supplied by the perturbation trajectory.

\begin{table}[ht]
\centering
\caption{AUROC for the full detector and the matched strict margin-only
control across ScientistQA, TriviaQA, and GSM8K. Each comparison uses the same
rows, labels, grouping, and folds; $\Delta_{\mathrm{M}}$ is Full minus
Margin-only.}
\label{tab:strict-detector-controls}
\small
\setlength{\tabcolsep}{6pt}
\begin{tabular}{llrrrr}
\toprule
Benchmark & Model & $n$ & Margin-only & Full & $\Delta_{\mathrm{M}}$ \\
\midrule
ScientistQA & Llama   & 1,077 & .653 & .908 & +.255 \\
            & Mistral &   621 & .525 & .824 & +.299 \\
            & Qwen    & 1,204 & .586 & .861 & +.275 \\
            & Macro average & -- & .588 & .864 & +.276 \\
\midrule
TriviaQA & Llama   & 1,000 & .7621 & .9492 & +.1871 \\
         & Mistral & 1,000 & .7379 & .9637 & +.2258 \\
         & Qwen    & 1,000 & .7344 & .9095 & +.1751 \\
         & Macro average & -- & .7448 & .9408 & +.1960 \\
\midrule
GSM8K & Llama   & 942 & .4989 & .7473 & +.2484 \\
      & Mistral & 942 & .6069 & .7439 & +.1370 \\
      & Qwen    & 942 & .6688 & .7603 & +.0915 \\
      & Macro average & -- & .5915 & .7505 & +.1590 \\
\bottomrule
\end{tabular}
\end{table}

\begin{table}[ht]
\centering
\caption{Additional ScientistQA metrics for the margin-only comparison in
Table~\ref{tab:strict-detector-controls}. TriviaQA and GSM8K are omitted here
because corresponding AUPRC, balanced-accuracy, and Macro-F1 summaries are not
reported.}
\label{tab:strict-detector-controls-additional}
\small
\setlength{\tabcolsep}{8pt}
\begin{tabular}{llrrr}
\toprule
Model & Input & AUPRC & Bal. acc. & Macro-F1 \\
\midrule
Llama & Full        & .928 & .827 & .823 \\
      & Margin-only & .721 & .604 & .598 \\
\midrule
Mistral & Full        & .845 & .737 & .734 \\
        & Margin-only & .535 & .529 & .527 \\
\midrule
Qwen & Full        & .871 & .776 & .775 \\
     & Margin-only & .596 & .564 & .563 \\
\bottomrule
\end{tabular}
\end{table}

On ScientistQA, the full detector improves AUROC over margin-only by
.255--.299 across the three backbones, with the same conclusion for AUPRC,
balanced accuracy, and Macro-F1. The result replicates in all six TriviaQA and
GSM8K settings: the gains range from .1751 to .2258 on TriviaQA and from .0915
to .2484 on GSM8K, for macro-average gains of .1960 and .1590, respectively.
The full detector therefore cannot be reduced to a nonlinear transformation
of the model's original answer margin; the response to targeted interventions
supplies substantial additional discriminative information.

\FloatBarrier

\subsection{Perturbation versus Non-Perturbation Features}
\label{app:perturb-vs-nonperturb}

This comparison asks whether perturbation-derived features add information
beyond hidden states extracted without an intervention. The non-perturbation
representation contains the unperturbed generated- and alternative-candidate
states together with the layer-14 answer representation. The matched
perturbation representation adds keyword-induced hidden-state deltas. Both use
identical folds within each matched comparison. ScientistQA uses
person-grouped five-fold OOF prediction repeated over seeds 42--44, and its
transfer experiment fits all preprocessing and the classifier on ScientistQA
before evaluating them unchanged on the multidomain target.

\begin{table}[ht]
\centering
\caption{Matched AUROC comparison of non-perturbation hidden states and the
same representation augmented with keyword perturbation hidden-state deltas.
ScientistQA uses person-grouped $3\times5$ OOF evaluation; multidomain results
use frozen transfer from ScientistQA. Gain is the augmented representation
minus the non-perturbation representation.}
\label{tab:perturb-vs-nonperturb-hidden}
\small
\setlength{\tabcolsep}{3pt}
\begin{tabular}{lrrrrrr}
\toprule
& \multicolumn{3}{c}{ScientistQA OOF} & \multicolumn{3}{c}{Multidomain transfer} \\
\cmidrule(lr){2-4}\cmidrule(lr){5-7}
Model & Non-perturb & + Perturb & Gain & Non-perturb & + Perturb & Gain \\
\midrule
Llama   & .894 & .910 & +.016 & .854 & .896 & +.042 \\
Mistral & .807 & .825 & +.018 & .695 & .832 & +.137 \\
Qwen    & .762 & .853 & +.091 & .627 & .829 & +.202 \\
\midrule
Macro average & .821 & .863 & +.091 & .725 & .852 & +.127 \\
\bottomrule
\end{tabular}
\end{table}

Table~\ref{tab:perturb-vs-nonperturb-hidden} provides the direct matched
comparison. Perturbation features yield in-domain improvements for
Llama and Mistral (.016 and .018), and a substantially larger .091 gain for
Qwen. The transfer pattern is stronger and fully consistent:
AUROC increases by .042, .137, and .202 for Llama, Mistral, and Qwen,
respectively, with a macro-average gain of .127 compared with .035 in-domain.
The perturbation-induced representation shift therefore contributes most
clearly under distribution shift, where unperturbed hidden states alone lose
considerably more predictive information.

\subsubsection{Compatibility with Aiersilan Representations}

We evaluate compatibility under two protocols. The first follows the official
Aiersilan MLP setup using random 70/10/20 train/validation/test splits for
seeds 42--44. We apply this paper-faithful protocol first on full ScientistQA
and then on TriviaQA and GSM8K. The second is the stricter person-grouped
ScientistQA cohort used elsewhere in this paper and treats Aiersilan hidden
states as a compatible feature block rather than reproducing its official MLP
protocol.

\begin{table}[ht]
\centering
\caption{Aiersilan--perturbation fusion under the paper-faithful MLP protocol
across ScientistQA, TriviaQA, and GSM8K. Values are mean $\pm$ standard
deviation over seeds 42--44. Fusion concatenates the Aiersilan and
perturbation representations.}
\label{tab:aiersilan-official-mlp-fusion}
\label{tab:aiersilan-official-mlp-fusion-trivia-gsm8k}
\small
\setlength{\tabcolsep}{3.5pt}
\begin{tabular}{lllrrr}
\toprule
Benchmark & Model & Metric & Aiersilan & Perturbation & Fusion \\
\midrule
ScientistQA & Llama & AUROC & $.818 $ & $.823$ & $.828$ \\
            & Mistral & AUROC & $.796$ & $.812$ & $.813$ \\
            & Qwen & AUROC & $.754$ & $.798$ &$.802$  \\
\midrule
Transfer & Llama & AUROC & $.781$ & $.824$ & $.865$ \\
        & Mistral & AUROC & $.682$ & $.840$ & $.856$ \\
        & Qwen & AUROC & $.771$ & $.822$ & $.829$ \\
\midrule
TriviaQA & Llama   & AUROC & $.8512$ & $.9361$ & $.9352$ \\
         & Mistral & AUROC & $.9282$ & $.9608$ & $.9582$ \\
         & Qwen    & AUROC & $.8542$ & $.9024$ & $.8660$ \\
\midrule
GSM8K & Llama   & AUROC & $.7501$ & $.7360$ & $.7368$ \\
      & Mistral & AUROC & $.7358$ & $.7364$ & $.7459$ \\
      & Qwen    & AUROC & $.7202$ & $.7373$ & $.7581$ \\
\bottomrule
\end{tabular}
\end{table}

Under the official MLP protocol, fusion improves over Aiersilan alone by .010
AUROC and .017 AUPRC. AUROC increases in each of the three seeds, so the
direction is consistent, although no paired significance test is available and
we therefore do not characterize the gain as statistically significant.

The additional benchmarks show that fusion is dataset-dependent. On TriviaQA,
perturbation alone outperforms Aiersilan for all three backbones, while fusion
is slightly below perturbation for Llama ($-.0009$) and Mistral ($-.0026$) and
more clearly below it for Qwen ($-.0364$). Fusion nevertheless remains above
Aiersilan alone in all three cases. On GSM8K, fusion improves over perturbation
alone by .0008, .0095, and .0208 for Llama, Mistral, and Qwen, respectively;
relative to Aiersilan alone, it is lower for Llama ($-.0133$) but higher for
Mistral (+.0101) and Qwen (+.0379). These results indicate complementary
signals on GSM8K, especially for Qwen, but also show that concatenating the two
representations does not guarantee an improvement when one component is
already strong.

The grouped-cohort results in
Table~\ref{tab:aiersilan-perturb-hidden-only-fusion} show a larger effect.
Adding perturbation hidden-state features raises ScientistQA AUROC from .7286
to .8616, a gain of 13.30 points. Under frozen multidomain transfer, AUROC
rises from .7577 to .9318, or 17.41 points. The transfer improvement is
positive in every target domain: 21.86 points for athletes, 14.81 for
musicians, and 17.22 for buildings. Because this grouped experiment uses an
Aiersilan-compatible hidden probe rather than the official MLP reproduction,
the two protocols should not be treated as numerically interchangeable. Taken
together, the results show that perturbation features often add information
beyond the unperturbed representation, particularly under distribution shift,
but that the benefit of direct feature-level fusion depends on the benchmark
and backbone.

\FloatBarrier

\subsection{Targeted Keyword versus Random-Span Perturbation}
\label{app:keyword-vs-random-ablation}

We next separate the value of performing an intervention from the value of
targeting the intervention at a selected keyword. The $2\times2$ design varies
the perturbation strategy used to construct the ScientistQA training features
and the strategy used at multidomain evaluation. It therefore distinguishes
in-domain separability from cross-domain alignment between the learned and
evaluated perturbation responses.

\subsubsection{Complete Detector Representation}

\begin{table}[ht]
\centering
\caption{The $2\times2$ transfer comparison using the complete detector
representation. The detector is trained on ScientistQA features produced by
keyword or random perturbations and evaluated on multidomain features produced
by either strategy. Transfer change is target AUROC minus source OOF AUROC.
Retention is the percentage of above-chance source AUROC retained on the
target, so values above 100\% indicate improvement.}
\label{tab:full-feature-perturbation-transfer-2x2}
\small
\setlength{\tabcolsep}{2pt}
\begin{tabular}{lllrrrr}
\toprule
Model & \shortstack{ScientistQA\\training} & \shortstack{Multidomain\\perturbation}
& \shortstack{ScientistQA\\OOF} & \shortstack{Multidomain\\AUROC}
& \shortstack{Transfer\\change $\uparrow$} & \shortstack{Retention\\$\uparrow$} \\
\midrule
Llama   & Keyword & Keyword & .908 & .933 & +.025 & 106.2\% \\
        & Keyword & Random  & .908 & .897 & $-.011$ & 97.3\% \\
        & Random  & Keyword & .889 & .920 & +.031 & 108.0\% \\
        & Random  & Random  & .889 & .883 & $-.006$ & 98.5\% \\
\midrule
Mistral & Keyword & Keyword & .824 & .843 & +.019 & 105.8\% \\
        & Keyword & Random  & .824 & .733 & $-.091$ & 71.9\% \\
        & Random  & Keyword & .815 & .789 & $-.027$ & 91.5\% \\
        & Random  & Random  & .815 & .733 & $-.082$ & 74.0\% \\
\midrule
Qwen    & Keyword & Keyword & .861 & .853 & $-.009$ & 97.6\% \\
        & Keyword & Random  & .861 & .724 & $-.137$ & 62.0\% \\
        & Random  & Keyword & .780 & .778 & $-.002$ & 99.2\% \\
        & Random  & Random  & .780 & .757 & $-.023$ & 91.8\% \\
\bottomrule
\end{tabular}
\end{table}

Random selection is itself a viable strategy for choosing intervention targets,
alongside exact, attention-based, and gradient-based selection. This explains
why random--random remains competitive: although randomly selected spans need
not be the most informative keywords, some still influence the model's answer,
and the complete detector also includes effective non-perturbation features.
Nevertheless, selecting targets carefully improves performance. The standard
exact keyword--keyword configuration outperforms random--random by .050 for
Llama, .110 for Mistral, and .096 for Qwen in transfer. Likewise, replacing
keyword targets with random targets at evaluation lowers the AUROC of a
keyword-trained detector by .036, .110, and .129, respectively. Thus random
selection provides a useful but weaker keyword-selection baseline, whereas
exact selection identifies more informative and more transferable targets.

\subsubsection{Perturbation-Induced Hidden-State Deltas Only}

\begin{table}[ht]
\centering
\caption{The $2\times2$ transfer comparison using only perturbation-induced
hidden-state delta features. Columns and transfer statistics follow
Table~\ref{tab:full-feature-perturbation-transfer-2x2}.}
\label{tab:perturb-hidden-delta-transfer-2x2}
\small
\setlength{\tabcolsep}{2pt}
\begin{tabular}{lllrrrr}
\toprule
Model & \shortstack{ScientistQA\\training} & \shortstack{Multidomain\\perturbation}
& \shortstack{ScientistQA\\OOF} & \shortstack{Multidomain\\AUROC}
& \shortstack{Transfer\\change $\uparrow$} & \shortstack{Retention\\$\uparrow$} \\
\midrule
Llama   & Keyword & Keyword & .783 & .859 & +.076 & 127.0\% \\
        & Keyword & Random  & .783 & .765 & $-.017$ & 93.9\% \\
        & Random  & Keyword & .725 & .864 & +.138 & 161.3\% \\
        & Random  & Random  & .725 & .784 & +.058 & 125.8\% \\
\midrule
Mistral & Keyword & Keyword & .679 & .832 & +.153 & 185.3\% \\
        & Keyword & Random  & .679 & .615 & $-.064$ & 64.1\% \\
        & Random  & Keyword & .600 & .752 & +.151 & 250.7\% \\
        & Random  & Random  & .600 & .601 & .000 & 100.4\% \\
\midrule
Qwen    & Keyword & Keyword & .755 & .831 & +.076 & 129.7\% \\
        & Keyword & Random  & .755 & .677 & $-.078$ & 69.4\% \\
        & Random  & Keyword & .603 & .707 & +.104 & 201.6\% \\
        & Random  & Random  & .603 & .633 & +.030 & 129.1\% \\
\bottomrule
\end{tabular}
\end{table}

With only perturbation-induced hidden-state deltas, the target perturbation
strategy becomes decisive. For keyword-trained detectors, replacing keyword
target perturbations with random ones lowers transfer AUROC by .094 for Llama,
.217 for Mistral, and .154 for Qwen. The same direction holds for
random-trained detectors: keyword target perturbations outperform random
target perturbations by .080, .151, and .074, respectively. This cross-direction
result is important because it does not depend on the training perturbations
being keywords; the hidden-state response elicited at selected keywords is
more informative on the new domains under this matched evaluation.

\begin{table}[ht]
\centering
\caption{In-domain AUROC using only perturbation-induced hidden-state
deltas on TriviaQA and GSM8K. Values are mean $\pm$ standard deviation over
seeds 42--44. Keyword advantage is Keyword minus Random.}
\label{tab:weighted-hidden-delta-keyword-random-trivia-gsm8k}
\small
\setlength{\tabcolsep}{6pt}
\begin{tabular}{llrrr}
\toprule
Benchmark & Model & Keyword & Random & Keyword advantage \\
\midrule
TriviaQA & Llama   & $.755 \pm .003$ & $.626 \pm .005$ & +.129 \\
         & Mistral & $.628 \pm .004$ & $.537 \pm .006$ & +.091 \\
         & Qwen    & $.659 \pm .011$ & $.507 \pm .006$ & +.152 \\
         & Macro average & .681 & .557 & +.124 \\
\midrule
GSM8K & Llama   & $.664 \pm .003$ & $.649 \pm .004$ & +.015 \\
      & Mistral & $.550 \pm .006$ & $.473 \pm .006$ & +.078 \\
      & Qwen    & $.499 \pm .016$ & $.473 \pm .009$ & +.026 \\
      & Macro average & .571 & .532 & +.039 \\
\bottomrule
\end{tabular}
\end{table}

\iffalse
\begin{table}[ht]
\centering
\caption{AUROC comparison of perturbation-induced hidden-state delta
aggregation under keyword and random-span perturbations. Keyword advantage is
the keyword AUROC minus the random AUROC.}
\label{tab:perturb-hidden-feature-inputs}
\small
\setlength{\tabcolsep}{4pt}
\begin{tabular}{llrrr}
\toprule
Model & Delta aggregation & Keyword AUROC & Random AUROC & Keyword advantage \\
\midrule
Llama   & Weighted & .783 & .725 & +.057 \\
Llama   & Mean     & .850 & .772 & +.078 \\
\midrule
Mistral & Weighted & .679 & .600 & +.079 \\
Mistral & Mean     & .673 & .598 & +.075 \\
\midrule
Qwen    & Weighted & .755 & .603 & +.153 \\
Qwen    & Mean     & .740 & .570 & +.170 \\
\midrule
Macro average & Weighted & .739 & .643 & +.096 \\
Macro average & Mean     & .754 & .647 & +.108 \\
\bottomrule
\end{tabular}
\end{table}

The in-domain aggregation comparison yields the same conclusion. Keyword
perturbation outperforms random perturbation for every model under both
weighted and mean aggregation, with advantages from .057 to .170. The
macro-average advantage is .096 for weighted deltas and .108 for mean deltas.
Together with the $2\times2$ transfer results, this argues against the explanation
that arbitrary hidden-state displacement is sufficient: which span is
perturbed determines both in-domain separability and cross-domain transfer.
\fi

\subsubsection{Cross-Benchmark Generalization from TriviaQA to DROP}

We further test whether the detector transfers across benchmarks rather than
only across entity domains within a shared construction. We fit all
preprocessing steps and the classifier on TriviaQA and evaluate them unchanged
on DROP; no DROP labels are used for model selection or fitting.

\begin{table}[ht]
\centering
\caption{Frozen benchmark transfer from TriviaQA to DROP. TriviaQA OOF AUROC
measures source-domain performance; DROP AUROC and AUPRC are computed after
applying the fitted detector without target-domain adaptation.}
\label{tab:triviaqa-drop-transfer}
\small
\setlength{\tabcolsep}{8pt}
\begin{tabular}{llrrr}
\toprule
Model & Selection & TriviaQA AUROC & DROP AUROC & DROP AUPRC \\
\midrule
Llama   & Exact     & .9492 & .7090 & .7316 \\
        & Attention & .9455 & .7069 & .7353 \\
\midrule
Mistral & Exact     & .9637 & .8523 & .8756 \\
        & Attention & .9636 & .8502 & .8651 \\
\midrule
Qwen    & Exact     & .9095 & .7052 & .7015 \\
        & Attention & .9021 & .7256 & .7175 \\
\midrule
Macro average & Exact     & .9408 & .7555 & .7696 \\
              & Attention & .9371 & .7609 & .7726 \\
\bottomrule
\end{tabular}
\end{table}

Table~\ref{tab:triviaqa-drop-transfer} shows that the learned perturbation
signal remains informative under a substantial shift from open-domain trivia
to reading comprehension. Every model--selection configuration transfers above
chance, with particularly strong performance for Mistral: exact selection
reaches .8523 AUROC and .8756 AUPRC, and attention selection closely matches it
at .8502/.8651. Qwen also transfers robustly, reaching .7052 AUROC with exact
selection and improving to .7256 with attention screening. Attention remains
within .0021 AUROC of exact selection for both Llama and Mistral while improving
Qwen by .0204. Consequently, its macro-average transfer AUROC is slightly
higher than exact selection (.7276 versus .7222), as is its AUPRC (.7393 versus
.7362), despite using a cheaper screening strategy. These results provide
evidence that the detector captures perturbation responses that generalize
across task formats, and that attention-based screening can preserve---and in
some cases improve---this cross-benchmark signal.

\FloatBarrier

\FloatBarrier

\section{Supplementary Material for Section~\ref{sec:category}}
\label{app:category}

\subsection{Operational Regime Assignment}
\label{app:regime-assignment}

This section specifies the empirical assignment underlying
Table~\ref{tab:unified-hallucination-types}. The unit of analysis is an item
whose original greedy answer is incorrect. No detector output---including P,
U, R, E, a supervised error probability, or a threshold selected from an
AUROC analysis---is used to assign a regime. Instead, assignment combines an
independent capability probe with $K=6$ stochastic generations from the
unchanged original prompt. Let $e_{\mathrm{orig}}\in\{0,\ldots,6\}$ be the
number of incorrect original-prompt generations.

All regime cutoffs are specified directly from the behavioral definitions of
the four categories; they are not learned by a regression model, fitted to the
reported category frequencies, or selected to optimize downstream detector
performance. In particular, the capability threshold separates an ambiguous
probe response (knowledge deficit) from a confident preference for the wrong
alternative (wrong knowledge) or confident access to the correct answer. The
repeat-generation thresholds then distinguish persistent errors under
available knowledge (context distraction) from sampling-sensitive errors
(unstable inference).

The implementation has a shared decision structure but two kinds of capability
probe. ScientistQA and TriviaQA provide a continuous likelihood-based
probability $p_{\mathrm{probe}}$. For these benchmarks, the frozen assignment
uses
\begin{align*}
\textsc{WrongKnowledge}&:\quad p_{\mathrm{probe}} < 0.25,\\
\textsc{KnowledgeDeficit}&:\quad 0.25 \le p_{\mathrm{probe}} \le 0.75,\\
\textsc{ContextDistract}&:\quad p_{\mathrm{probe}} > 0.75
 \ \wedge\ e_{\mathrm{orig}}\in\{5,6\},\\
\textsc{UnstableInference}&:\quad p_{\mathrm{probe}} > 0.75
 \ \wedge\ e_{\mathrm{orig}}\in\{2,3,4\}.
\end{align*}
An otherwise probe-capable item with $e_{\mathrm{orig}}\in\{0,1\}$ is
\textsc{Unclassified}, as are missing or unverifiable probes.

GSM8K and DROP instead estimate capability by directly sampling the probe six
times. Their implementations therefore use integer success thresholds rather
than the continuous cutoffs above. A probe-capable item is
\textsc{ContextDistract} when $e_{\mathrm{orig}}\in\{5,6\}$ and
\textsc{UnstableInference} when
$e_{\mathrm{orig}}\in\{2,3,4\}$. If the capability probe fails, the item is
\textsc{WrongKnowledge} when one valid wrong probe answer is modal in at least
five of six samples and \textsc{KnowledgeDeficit} otherwise. The pass threshold
is four correct probe samples for GSM8K and five for DROP. These
benchmark-specific thresholds are important: treating all four probes as a
single $p_{\mathrm{probe}}>.75$ rule would not reproduce the reported counts.

The thresholds have a direct behavioral interpretation. For the binary
likelihood probes, $0.5$ corresponds to no preference between the correct and
incorrect alternatives. We therefore reserve the symmetric interval
$[0.25,0.75]$ for ambiguous evidence: a model lacking the relevant knowledge
should not reliably know which alternative to select and should remain near
chance. Values below $0.25$ instead indicate a pronounced preference for the
incorrect alternative, which operationalizes a stable wrong belief, whereas
values above $0.75$ provide positive evidence that the required knowledge is
available. Conditional on such availability, five or six errors among the six
original-prompt samples indicate a persistent failure and are labeled
\textsc{ContextDistract}; two to four errors indicate sensitivity to sampling
and are labeled \textsc{UnstableInference}. Zero or one error is left
unclassified because the initially incorrect greedy response is not reproduced
often enough to establish either failure mode. The sampled probes use the same
logic in discrete form: repeated selection of one identical wrong answer in at
least five of six trials suggests wrong knowledge, while dispersed failures
are more consistent with a knowledge deficit. The four-of-six GSM8K and
five-of-six DROP capability cutoffs reflect the reliability criteria fixed for
their different probe constructions, rather than post-hoc optimization on
detector performance.

The resulting labels are mutually exclusive operational signatures, not
identified latent causes.

\paragraph{ScientistQA probe.}
The probe consists of closed-book binary questions about the discriminative
facts associated with the two candidate scientists; the original profiles are
not supplied. Each fact is queried for both candidate owners. To remove the
model's general tendency to answer ``Yes'' to both candidates, we form a
paired owner probability. If $g$ is the gold owner and $d$ the other owner for
fact $f$, then
\[
p_f=\sigma\!\left(\operatorname{logit}p(\mathrm{Yes}\mid g,f)
-\operatorname{logit}p(\mathrm{Yes}\mid d,f)\right),\qquad
p_{\mathrm{probe}}=|F|^{-1}\sum_{f\in F}p_f.
\]
This paired construction is important: averaging the two binary accuracies
directly would turn a ``Yes/Yes'' response into an artificial probability near
$0.5$. The Full cohort contains all aligned erroneous ScientistQA items. The
Known cohort is the pre-existing experiment-88 subset with
\texttt{probe\_known=True}; it is not reselected using the continuous
probability above. Consequently, some items in this legacy Known cohort fall
in the new continuous \textsc{KnowledgeDeficit} interval.

The label ``Scientist'' in Table~\ref{tab:empirical-regime-prevalence}
means the full ScientistQA sample cohort under the names/short-description
prompt. It must not be confused with the separate condition that supplies both
complete biographical profiles. The complete-profile condition has not yet
been assigned regimes and is not included in the table. The full and Known
columns contain 1,334 and 453 original errors, respectively.

\paragraph{TriviaQA probe.}
TriviaQA uses a closed-book question without the retrieved context. The cached
correct-versus-wrong answer log-likelihood margin $m$ is converted to
$p_{\mathrm{probe}}=\sigma(m)$. Answer correctness in the six original-prompt
generations is evaluated after TriviaQA alias normalization. Thus a stable
error under the retrieved context is called context distraction only when the
closed-book probe strongly supports the correct answer; stable error alone is
not treated as evidence of wrong knowledge.

\paragraph{GSM8K probe.}
We construct a structured computation probe from the arithmetic expressions in
the gold rationale. Specifically, we extract the ordered
\texttt{<<expression=result>>} chain, remove each result, and omit the original
word-problem story and gold final answer. The model completes this expression
chain six times, and $p_{\mathrm{probe}}$ is the fraction of generations that
produce the correct final numeric answer. Regimes are then assigned using the
sample-based rules above.

\paragraph{DROP probe.}
Because DROP provides neither gold programs nor supporting-sentence
annotations, we construct a minimal-evidence probe by ranking passage sentences
according to lexical overlap with the question and selecting fixed top-one or
top-two evidence sets. We then test a question-conditioned operation inventory
covering sums, absolute differences, percentage complements, ratios, direct
numeric lookup, and extractive selection. The gold answer is used only to
verify that the selected evidence and a unique compatible operation recover
the annotation; model predictions and detector scores are never used during
construction. Span-answer probes require two evidence sentences so that the
reduced prompt does not reveal only the answer entity. The resulting probes are
evaluated using the same six-sample rules above.

\subsection{Prevalence Results}
\label{app:regime-prevalence}

\begin{table}[ht]
\centering
\caption{Empirical regime assignment among originally incorrect responses.
Counts are followed by percentages within each benchmark. Scientist$^\star$
denotes the subset of ScientistQA for which the relevant knowledge is
behaviorally verified as available to the model. The four regimes are mutually
exclusive.}
\label{tab:empirical-regime-prevalence}
\small
\setlength{\tabcolsep}{2.5pt}
\newcommand{\regimecell}[2]{#1\,{\scriptsize (#2\%)}}
\begin{tabular}{lccccc}
\toprule
Category & \shortstack{Scientist\\($n=1334$)} &
\shortstack{Scientist$^\star$\\($n=453$)} &
\shortstack{TriviaQA\\($n=500$)} & \shortstack{GSM8K\\($n=471$)} &
\shortstack{DROP\\($n=500$)} \\
\midrule
Wrong knowledge & \regimecell{72}{5.40} & \regimecell{0}{0.00}
& \regimecell{238}{47.60} & \regimecell{34}{7.20} & \regimecell{43}{8.60} \\
Knowledge deficit & \regimecell{573}{42.95} & \regimecell{66}{14.57}
& \regimecell{187}{37.40} & \regimecell{99}{21.00} & \regimecell{120}{24.00} \\
Context distract & \regimecell{463}{34.71} & \regimecell{258}{56.95}
& \regimecell{56}{11.20} & \regimecell{222}{47.10} & \regimecell{285}{57.00} \\
Unstable inference & \regimecell{204}{15.29} & \regimecell{118}{26.05}
& \regimecell{12}{2.40} & \regimecell{94}{20.00} & \regimecell{49}{9.80} \\
Unclassified & \regimecell{22}{1.65} & \regimecell{11}{2.43}
& \regimecell{7}{1.40} & \regimecell{22}{4.70} & \regimecell{3}{0.60} \\
\bottomrule
\end{tabular}
\end{table}

\paragraph{Interpretation.}
The within-benchmark shifts, rather than a literal cross-benchmark comparison
of percentages, provide the cleanest evidence. Restricting ScientistQA to the
legacy probe-known cohort reduces knowledge deficit from $42.95\%$ to
$14.57\%$ and increases context distraction from $34.71\%$ to $56.95\%$.
GSM8K contains both a large context-distraction component ($47.10\%$) and a
combined $41.00\%$ knowledge-deficit/unstable component, which motivates the
fusion experiments below. TriviaQA is instead dominated by failed closed-book
probes ($85.00\%$ across wrong knowledge and knowledge deficit).

These are behavioral signatures rather than identified latent causes.
ScientistQA and TriviaQA use likelihood-derived probabilities, GSM8K uses a
gold-rationale computation probe, and DROP combines an automatic validator
with manual probe auditing. Their capability measurements are therefore not
calibrated to a common scale. In particular, the DROP percentages should not
be interpreted as the output of a fully automatic taxonomy. Unclassified
items indicate that the operational rules do not support a unique assignment,
not that the error lacks a systematic cause.

\subsection{Method Fusion}
\label{app:fusion}
\subsubsection{Perturbation and Representation}

GSM8K contains substantial mass in both context-distraction and
knowledge-deficit/unstable-inference errors, so it provides a direct test of
whether perturbation and representation signals cover complementary failure
patterns. We combine the exact perturbation detector (P) with the
Aiersilan-style representation probe (R) using nested stacking. For each
ordinary stratified 80/20 split, both base detectors are fitted on the
development partition, and a logistic stacker is trained only on their
five-fold out-of-fold development predictions before evaluation on the
untouched test partition. We repeat this procedure for seeds 42--47, fitting
all preprocessing within the corresponding training data.

Table~\ref{tab:pu-fusion-gsm8k} shows that P+R improves over the stronger
individual detector for every backbone. The AUROC gains are .011 for Llama,
.014 for Mistral, and .004 for Qwen; the corresponding AUPRC gains are .012,
.022, and .001. These gains are modest and vary across models, indicating that
the two signals overlap substantially, but their consistent direction supports
the predicted complementarity between localized answer dependence and
response-level representations.

  \begin{table}[h]
  \centering
  \caption{Nested perturbation--representation fusion under ordinary
  stratified 80/20 in-domain splits on GSM8K. Results are averaged over
  seeds 42--47.}
  \label{tab:pu-fusion-gsm8k}
  \small
  \setlength{\tabcolsep}{8pt}
  \begin{tabular}{llrr}
  \toprule
  Backbone & Method & AUROC & AUPRC \\
  \midrule
  Llama
      & Perturbation (Exact)              & .732 & .751 \\
      & Representation (Aiersilan-style)  & .736 & .758 \\
      & P+R                               & \textbf{.747} & \textbf{.770} \\
  \midrule
  Mistral
      & Perturbation (Exact)              & .747 & .763 \\
      & Representation (Aiersilan-style)  & .696 & .729 \\
      & P+R                               & \textbf{.761} & \textbf{.785} \\
  \midrule
  Qwen
      & Perturbation (Exact)              & .740 & .761 \\
      & Representation (Aiersilan-style)  & .720 & .740 \\
      & P+R                               & \textbf{.744} & \textbf{.762} \\
  \bottomrule
  \end{tabular}
  \end{table}

\FloatBarrier

\subsubsection{Perturbation and Uncertainty}
\label{app:selfcheck-fusion}
We next combine perturbation with SelfCheckGPT (U) on GSM8K, where the regime
distribution predicts that local answer dependence and sampling uncertainty
should provide complementary signals. SelfCheckGPT draws 20 samples at
temperature $1.0$ and top-$p=1.0$ with generation seed 20260823, then averages
NLI contradiction probabilities over sampled passages and primary-response
sentences. For each ordinary stratified 80/20 outer split, P is fitted on the
development partition. A logistic stacker is trained on five-fold OOF scores
within that partition and evaluated once on the untouched outer test
partition. The outer seeds are 42--47. Because all three scoring backbones use
the same frozen GSM8K generations and SelfCheckGPT does not depend on the
scoring backbone, the U-only result is identical across the three blocks.

\begin{table}[h]
  \centering
  \caption{Nested perturbation--uncertainty fusion under stratified
  80/20 in-domain splits on GSM8K. Results are averaged over seeds 42--47.}
  \label{tab:pu-selfcheck-fusion-gsm8k}
  \small
  \setlength{\tabcolsep}{8pt}
  \begin{tabular}{llrr}
  \toprule
  Backbone & Method & AUROC & AUPRC \\
  \midrule
  Llama
      & Perturbation (Exact)       & .732 & .778 \\
      & Uncertainty (SelfCheckGPT) & .783 & .762 \\
      & P+U                        & \textbf{.821} & \textbf{.813} \\
  \midrule
  Mistral
      & Perturbation (Exact)       & .759 & .770 \\
      & Uncertainty (SelfCheckGPT) & .783 & .762 \\
      & P+U                        & \textbf{.821} & \textbf{.806} \\
  \midrule
  Qwen
      & Perturbation (Exact)       & .760 & .774 \\
      & Uncertainty (SelfCheckGPT) & .783 & .762 \\
      & P+U                        & \textbf{.821} & \textbf{.812} \\
  \bottomrule
  \end{tabular}
  \end{table}

Table~\ref{tab:pu-selfcheck-fusion-gsm8k} shows a consistent gain across
backbones. P+U reaches .821 AUROC in all three cases, improving over U by .038
and over P by .059, .062, and .061 for Llama, Mistral, and Qwen, respectively.
For AUPRC, fusion improves over the stronger individual detector---P in all
three blocks---by .035, .036, and .038. The agreement across scoring backbones
supports the predicted complementarity between localized perturbation effects
and generation-level uncertainty on GSM8K. It does not imply that fusion is
universally beneficial: a separate ScientistQA diagnostic produces no gain,
indicating that complementarity depends on the benchmark's mixture of failure
regimes. Those ScientistQA runs use ordinary rather than person-grouped splits
and therefore do not replace the main group-disjoint metrics. No directly
comparable completed nested P+U run is available for TriviaQA or DROP.
\FloatBarrier

\FloatBarrier
% !TeX root = iclr2027_conference.tex
\section{Limitations}
\label{app:limitations}

Our study focuses on hallucination settings where model errors can be meaningfully analyzed 
through input perturbations, and we evaluate the proposed framework on a representative but 
necessarily limited set of benchmarks and model families. While the results are consistent 
across multiple domains, they do not establish that the same perturbation patterns will 
hold for every model architecture, training distribution, or type of hallucination. In 
particular, hallucinations arising from substantially different mechanisms may require 
additional signals beyond those considered in this work.

Second, the full implementation is currently most directly applicable to open-weight models because its
contrastive scoring uses token-level likelihoods and some detector variants use internal representations. A
restricted perturbation-only variant may nevertheless be possible in black-box settings. For example, one could
issue the original query together with targeted keyword deletions or replacements and length- and position-
matched control edits, then measure changes in answer choice, response consistency, or externally verified
correctness across repeated queries. Comparing targeted edits with matched controls would preserve the central
intervention logic while replacing likelihood differences with response-level effects. Such a variant would
require additional queries and would provide a coarser signal than the white-box method; evaluating whether it
retains comparable detection performance on closed-source frontier models remains future work.

\end{document}